\pdfoutput=1
\documentclass[10pt,twocolumn,letterpaper]{article}

\usepackage[T1]{fontenc}
\usepackage{times}
\usepackage[letterpaper,margin=0.75in,columnsep=0.3in]{geometry}
\usepackage{amsmath,amssymb}
\usepackage{graphicx}
\usepackage{booktabs}
\usepackage{array}
\usepackage[table]{xcolor}
\usepackage{enumitem}
\usepackage[hidelinks]{hyperref}
\hypersetup{pdftitle={PARMA: PARallel Memory Arbitration Policies for SAM-Based Visual Object Tracking}, pdfauthor={Yitong Wang, Haoyang Tang, Jiayi Jiang, Mohit Gupta, Hongyi Wang}}
\usepackage{url}
\usepackage{capt-of}

\setlist{nosep,leftmargin=*}
\newcommand{\se}[1]{\,(#1)}
\newcommand{\ReadU}{\textsc{U}}
\definecolor{upgreen}{HTML}{2E7D32}
\definecolor{downred}{HTML}{C62828}
\newcommand{\up}[1]{\textcolor{upgreen}{#1}}
\newcommand{\down}[1]{\textcolor{downred}{#1}}
\newcommand{\gain}[1]{{\scriptsize\up{#1}}}
\newcommand{\gainsm}[1]{{\tiny\up{#1}}}
\newcommand{\loss}[1]{{\scriptsize\down{#1}}}
\newcommand{\second}[1]{\underline{#1}}
\newcommand{\cm}{\checkmark}

\title{PARMA: PARallel Memory Arbitration Policies for SAM-Based Visual Object Tracking}

\author{Yitong Wang$^{1}$ \quad Haoyang Tang$^{1}$ \quad Jiayi Jiang$^{1}$ \quad Mohit Gupta$^{1}$ \quad Hongyi Wang$^{2,*}$\\
$^{1}$University of Wisconsin--Madison \qquad $^{2}$Rutgers University\\
{\tt\small \{wang2895, htang96, jjiang375\}@wisc.edu, mohitg@cs.wisc.edu, hw689@cs.rutgers.edu}}
\date{}

\begin{document}
\twocolumn[{%
\renewcommand\twocolumn[1][]{#1}%
\maketitle
\vspace{-1.2em}
\begin{center}
\includegraphics[width=\textwidth]{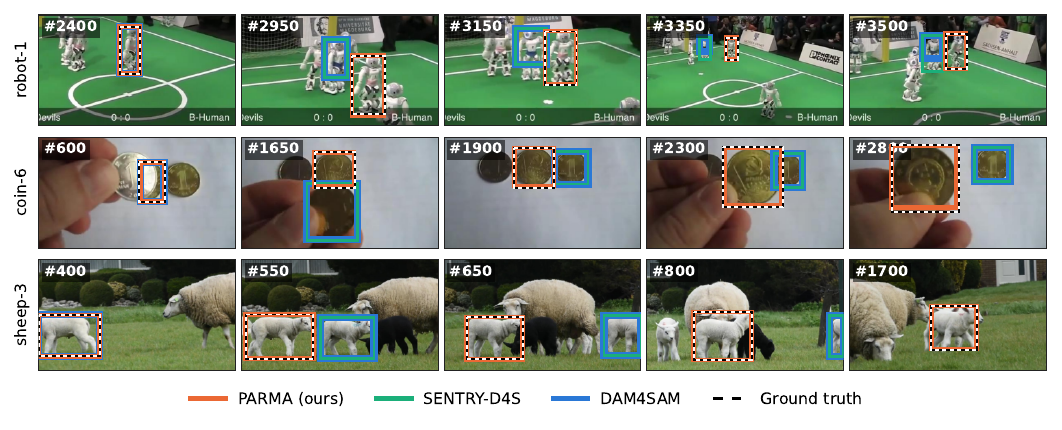}
\captionof{figure}{Three LaSOT test videos with look-alike distractors. PARMA (orange) against DAM4SAM (blue), its base tracker, and SENTRY-D4S (green), the strongest re-run baseline on LaSOT; dashed: ground truth. All trackers start on the target (first column). robot-1: from about frame 2{,}700 all five re-run baselines follow another robot; PARMA is back on the target by frame 2{,}800 and stays on it in the frames shown. coin-6: DAM4SAM and SENTRY-D4S move to the other coin (SAMURAI also loses the target). sheep-3: DAM4SAM and SENTRY-D4S jump to a neighbouring lamb at frame 498. Frame numbers start from 0.}
\label{fig:teaser}
\end{center}
\vspace{0.6em}
}]
{\renewcommand{\thefootnote}{\fnsymbol{footnote}}\footnotetext[1]{Corresponding author.}}

\begin{abstract}
Memory-based visual object trackers must adapt to appearance changes while preventing erroneous predictions from contaminating subsequent tracking. Different memory policies make different trade-offs between these objectives, creating complementary strengths that a tracker can exploit. Inspired by mixture of experts, we propose PARMA (PARallel Memory Arbitration), a framework that treats parallel memory-policy branches as tracking experts and learns which output to trust at each frame. The experts share a frozen Segment Anything~2.1 model and image features while maintaining distinct tracking histories. A learned gate uses current and past branch signals to select among competing predictions, without modifying any branch's memory. With common box extraction and scoring, PARMA achieves SOTA results among the SAM\,2.1-L trackers in our controlled evaluation. It outperforms all five re-run baselines on all three metrics across LaSOT, LaSOT-ext, UAV123 and NFS, reaching 77.20 success AUC on LaSOT and 63.30 on LaSOT-ext. On TNL2K, an external benchmark unused during development, PARMA reaches 62.73 success AUC on 699 test sequences. Pre-registered tests show gains of 1.89 AUC points over the DAM4SAM branch of the same run and 1.32 over an untrained consensus rule whose switching rate was matched on training data.
\end{abstract}

\section{Introduction}

Visual object tracking requires maintaining the identity of a target across a video despite changes in appearance, occlusion and visually similar distractors. For a tracker that uses past predictions as memory, this raises a central question: how can it adapt to new observations without allowing an incorrect prediction to corrupt future tracking? Segment Anything~2 (SAM\,2)~\cite{sam2} provides a strong foundation for this setting by conditioning each new frame on a streaming memory of the prompted frame and recent predictions. However, its performance depends on which observations are retained and retrieved. Memory selection therefore determines not only what the tracker remembers, but also how earlier errors influence later predictions.

Existing SAM\,2-based trackers address this problem from several directions. SAMURAI~\cite{samurai} filters memory using predicted mask quality, object confidence and motion consistency, while DAM4SAM~\cite{dam4sam} retains reliable observations containing distractors in a long-term memory. HiM2SAM~\cite{him2sam} combines hierarchical motion estimation with short- and long-term memory, and SAMITE~\cite{samite} refines memory admission and introduces position prompts. Beyond selecting individual frames, SAM2Long~\cite{sam2long} maintains alternative memory paths and selects among them using accumulated scores. SENTRY~\cite{sentry} instead supplements a host tracker with candidate generation and verification. More recently, SAM2RL~\cite{sam2rl} and SAM3RL~\cite{sam3rl} have learned memory replacement decisions. These approaches improve memory management or prediction verification, but leave open whether the complementary outputs of different memory policies can be exploited during tracking. This perspective is consistent with recent evidence that short-term memory selection accounts for much of the difference between SAM\,2 trackers~\cite{rethinkmem}.

Our approach, PARMA (PARallel Memory Arbitration), is inspired by the mixture-of-experts (MoE) principle: a learned gating mechanism determines which experts to trust for a given input~\cite{jacobs1991moe}. We apply this idea to memory-based tracking by treating branches with different memory policies as experts and learning a causal gate over their outputs. Expert diversity comes from the policies and their resulting memory histories, while the underlying model weights remain shared and frozen. Restrictive memory admission can resist distractors but delay adaptation to appearance changes, whereas frequent updates can adapt quickly while retaining erroneous predictions. These trade-offs suggest that the most reliable expert may change over the course of a video. On 583 LaSOT training sequences, choosing the best cluster of agreeing candidates in each frame using ground truth would add 3.95 success AUC points over DAM4SAM. We train an arbiter to choose among these clusters using only signals available by the current frame. Fig.~\ref{fig:teaser} shows three LaSOT videos in which DAM4SAM and SENTRY-D4S follow a look-alike while PARMA stays on the target in the frames shown.

\begin{figure*}[t]
\centering
\includegraphics[width=\textwidth]{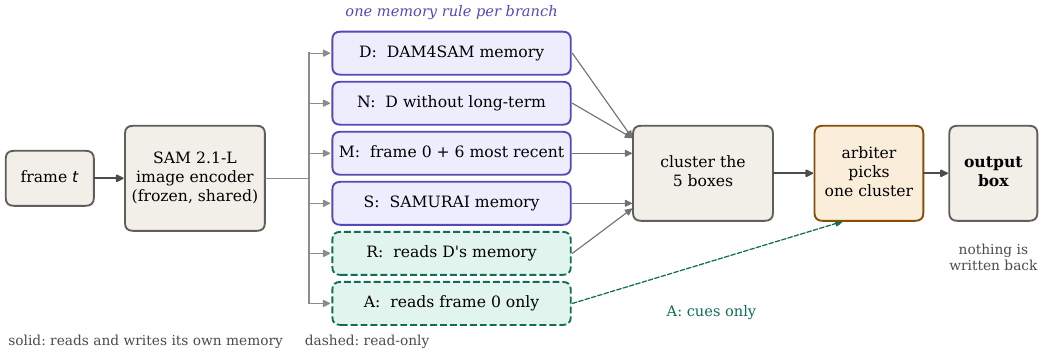}
\caption{Overview. One frozen SAM\,2.1-L image encoder runs once per frame. Each branch decodes a mask from the memory it reads: D, N, M and S read and write their own banks, R reads the entries of D that D and S agreed on, and A reads only the prompted frame. On frames where the five candidate boxes split into more than one cluster, the arbiter scores each cluster and reports one. A contributes cues only, and nothing is written back into any memory.}
\label{fig:method}
\end{figure*}

One frozen SAM\,2.1-L image encoder computes features once per frame, which are shared across six branches with different memory rules (Fig.~\ref{fig:method}, Table~\ref{tab:branches}). Five branches propose boxes and a sixth provides a reference to the initial target. The arbiter chooses which prediction to report. This choice leaves each branch's memory unchanged.

At each frame, we group the five candidate outputs into clusters according to box agreement. When the candidates form a single cluster, we report DAM4SAM's output. Otherwise, a small gradient-boosted regressor scores the clusters using 95 cues describing geometry, confidence, agreement and appearance, all derived from branch outputs and internal signals available up to the current frame. A competing cluster replaces DAM4SAM's cluster only when its predicted score exceeds the latter by a margin. The regressor is fitted on records from 350 LaSOT training sequences, with no fine-tuning of SAM\,2.1. In this MoE-inspired formulation, all branches continue to run, and the learned gate selects one representative output rather than averaging predictions; only the arbiter is trained. Selecting among trackers has substantial precedent~\cite{mcct,ltmu,aaa,cocolot,bofn}; our contribution is the combination of shared image features, alternative memory policies on one frozen model, and causal selection at the output alone. This separation also makes the arbiter's decisions reproducible from the recorded branch histories.

We evaluate this design through controlled comparisons that account for differences in box extraction and scoring across published trackers. We re-run SAM\,2.1, SAMURAI, HiM2SAM, DAM4SAM and SENTRY-D4S, extract boxes consistently from their archived masks, and score all systems with the same implementation of the official LaSOT protocol. PARMA reaches 77.20 AUC on LaSOT and 63.30 on LaSOT-ext, exceeding the strongest re-run baseline by 1.62 and 1.92 points, respectively. The LaSOT result is descriptive because the main configuration was selected after evaluation. Pre-registered comparisons against the DAM4SAM branch from the same run and archived SENTRY-D4S establish gains on LaSOT-ext and on the equally weighted pooled evaluation of LaSOT-ext, UAV123 and NFS, but not on UAV123 or NFS individually. The separate comparison with the strongest archived DAM4SAM baseline on LaSOT-ext is descriptive. Shared encoding limits the cost of running the parallel branches to approximately twice DAM4SAM's per-frame time in same-job measurements; this measurement excludes the CPU arbiter (Sec.~\ref{sec:speed}).

Because these benchmarks had all informed development in some way, we additionally conduct a single external evaluation on TNL2K~\cite{tnl2k}, which played no part in the method's design. Before running the tracker, we register comparisons against both the DAM4SAM branch from the same run and an untrained consensus rule whose switching rate is matched to the arbiter's on training data. The arbiter improves AUC by 1.89\se{0.26} and 1.32\se{0.21}, respectively, where parentheses denote paired per-sequence standard errors; both comparisons satisfy the pre-registered criteria after multiple-comparison correction. These results support the value of learning which memory-policy output to trust, beyond the gains provided by a conservative consensus rule.

\paragraph{Contributions.}
\begin{itemize}
\item \textbf{Parallel memory policies with learned output arbitration.} We run several memory policies on one frozen SAM\,2.1-L model and learn which prediction to report when the policies disagree, at about $2\times$ DAM4SAM's per-frame time; a two-candidate version keeps 78\% of the LaSOT gain at $1.21\times$.
\item \textbf{A controlled evaluation.} We re-run five SAM\,2.1-L trackers with common box extraction and scoring, and test the arbiter in a pre-registered external test on TNL2K, against a fixed branch and against an untrained consensus rule calibrated to its switching rate.
\item \textbf{Results.} PARMA sets a new state of the art among SAM\,2-based trackers on LaSOT (77.2 AUC) and LaSOT-ext (63.3) and ranks first in all 12 columns (AUC, $P_\text{norm}$ and $P$ on four benchmarks) of Table~\ref{tab:mainfull} under one evaluation.
\end{itemize}

\section{Related Work}

\paragraph{Memory design in segmentation-based tracking.}
SAM\,2~\cite{sam2} provides a promptable segmentation model whose streaming memory allows predictions to depend on earlier observations. This mechanism supports visual object tracking, but also creates a feedback loop: an inaccurate prediction can enter memory and influence subsequent localization. Several training-free extensions address this problem through more selective memory management. SAMURAI~\cite{samurai} combines mask-quality estimates with Kalman-based motion consistency to guide prediction and memory selection. DAM4SAM~\cite{dam4sam} separates recent appearance information from long-term observations that help distinguish the target from distractors, using an introspective update strategy to retain reliable and informative frames. HiM2SAM~\cite{him2sam} combines hierarchical motion estimation with short- and long-term memory optimization, while SAMITE~\cite{samite} evaluates the feature-wise and position-wise reliability of stored predictions and introduces positional prompts to reduce distraction. These methods make different choices about reliability, recency and adaptation. Alansari~et~al.~\cite{rethinkmem} study these choices across SAM\,2 and SAM\,3~\cite{sam3}, identifying short-term memory selection as a major source of differences and proposing hybrids that combine short-term appearance and long-term distractor-resolving memory. This work supports treating memory policy as an explicit design variable when constructing complementary tracking experts.

\paragraph{Multiple memory paths and learned memory control.}
SAM2Long~\cite{sam2long} addresses error accumulation by maintaining several segmentation pathways in a constrained memory tree. Candidate continuations are pruned using cumulative scores, and the final result is selected after the complete video has been processed. It therefore already explores multiple memory histories; the distinction from PARMA concerns how those histories are generated and how their outputs are selected. We run predefined memory policies in parallel and report a causal decision for each frame. A separate line learns the memory-management policy itself. The original SAM2RL study~\cite{sam2rl} formulates memory control as a reinforcement-learning problem and demonstrates its potential in a per-video overfitting setting. SAM3RL~\cite{sam3rl} introduces an RL-trained memory controller for SAM\,3 and also evaluates the approach on SAM\,2. Such controllers act on the observations available to future predictions. Our learned component instead acts on the outputs of existing policies and does not modify their memory updates. The approaches are compatible in principle: a learned memory controller could supply an expert within our framework, although we do not evaluate that combination here.

\paragraph{Tracker ensembles and adaptive selection.}
Combining trackers with complementary strengths is an established approach to visual tracking. HMMTxD~\cite{hmmtxd} models tracker failures as latent states in a hidden Markov model and combines tracker observations with information from an online detector. HDT~\cite{hdt} uses an adaptive Hedge mechanism to combine trackers constructed from different convolutional feature layers. MCCT~\cite{mcct} maintains feature-based experts, selects an expert using a robustness criterion, and uses disagreement to assess tracking reliability and guide updates. AAA~\cite{aaa} develops adaptive aggregation of arbitrary online trackers with a theoretical performance guarantee. These methods establish the relevance of expert diversity and reliability estimation; selecting among trackers is not itself our novelty. Learned selection is also well represented. DTNet~\cite{dtnet} uses hierarchical reinforcement learning to switch between detection-based and template-based trackers. Tracking-by-Trackers~\cite{tbt} combines knowledge distillation and reinforcement learning to train a student that supports standalone tracking, online adaptation and tracker fusion. BofN~\cite{bofn} learns to predict the best tracker from a pool using visual input, with both a video-level variant based on initial frames and a frame-level variant that repeats selection at temporal intervals. Our comparison with these methods therefore rests on the construction of the candidate experts and the information used for selection, rather than on the presence of a learned or time-varying selector alone.

\paragraph{Verification and control of tracker updates.}
Reliability estimates can also determine when to correct a tracker or update its model. PTAV~\cite{ptav} couples a fast tracker with a verifier that checks predictions and supplies corrections when needed. CoCoLoT~\cite{cocolot} combines complementary trackers through a learned verification model and a decision policy that selects a tracker and corrects failing ones. LTMU~\cite{ltmu} learns when an appearance model should be updated by integrating geometric, discriminative and appearance information over time. In SAM\,2-based tracking, SENTRY~\cite{sentry} verifies candidate predictions through backward tracklets and temporal consistency before admitting corrections to memory. Its released implementation preserves the host memory by default but can replace spatial memory for accepted rescue frames. Thus, verification can affect the subsequent tracking state in these systems. Our arbiter uses reliability cues to choose the reported output, while each branch continues under its prescribed memory policy regardless of that choice.

\paragraph{Mixture of experts and output gating.}
Our conceptual motivation comes from mixture of experts (MoE), where a learned gate assigns inputs to experts with complementary capabilities~\cite{jacobs1991moe}. Sparse MoE architectures extend this principle to conditional computation: Shazeer~et~al.~\cite{shazeer2017moe} activate a small subset of expert networks for each input, and Switch Transformers~\cite{fedus2022switch} simplify routing to a single expert per token. These architectures use distinct trainable expert parameters to increase model capacity while limiting active computation. Our experts instead share frozen SAM\,2.1 weights and obtain diversity primarily through different memory policies and histories. The gate is trained separately from the backbone and operates on already computed branch signals. It selects one representative prediction from clusters of agreeing candidates, without averaging their masks or feeding the selection back into their memories. The branches share one image encoding per frame, so each added branch costs only its own memory attention and mask decoding, plus a memory encoding if it writes memory; in our implementation the branches run one after another, and the full system takes about twice DAM4SAM's per-frame time (Sec.~\ref{sec:speed}). We borrow the idea of a learned gate from mixture-of-experts models~\cite{jacobs1991moe}. Here the gate selects among predictions already computed by branches that share model weights and use different memory rules.

\section{Method}
\label{sec:method}

PARMA takes in a video with frames $I_0,\dots,I_T$ and a target box $B_0$ on frame~0, and outputs, for every frame $t>0$, either a box $\hat B_t$ or an empty prediction $\varnothing$. The architecture consists of six memory branches on one shared SAM\,2.1 model and an arbiter that chooses among their predictions (Fig.~\ref{fig:method}). We train only the arbiter. Sections~\ref{sec:shared}--\ref{sec:arbiter} describe the shared model, the branches, the grouping of their predictions, and the arbiter.

\subsection{Shared encoder and per-branch decoding}
\label{sec:shared}

We use a frozen SAM\,2.1-L model~\cite{sam2} in the DAM4SAM runtime~\cite{dam4sam}. On frame~0 we report $B_0$, and DAM4SAM converts it into a target mask. The resulting memory entry is the \emph{anchor} shared by all branches. On each later frame $t$, the Hiera image encoder computes the features $F_t$ once. Each active branch then decodes a mask from the shared features and the memory entries specified by its read rule (Table~\ref{tab:branches}). A branch that stores memories also runs the memory encoder. All branches use the same weights.

\paragraph{Memory entries and reading.}
The entry of branch $k$ for frame $\tau$ holds a spatial memory, obtained by encoding $F_\tau$ together with the mask that $k$ selected on $\tau$, and an object pointer projected from the decoder token of that mask. Two branches that store the same frame therefore hold different entries whenever they selected different masks. To predict frame $t$, a branch concatenates its selected spatial memories (at most seven, including the anchor) and object pointers into one token sequence with spatial and temporal positional encodings, and $F_t$ cross-attends to this sequence. Entries selected within a branch are read jointly. Memories of different branches are never combined.

\paragraph{Decoding and box read-out.}
The mask decoder returns three mask hypotheses with predicted IoUs $q_1,q_2,q_3$ and an object-presence logit $o$. Under the native rule the target is absent when $o\le 0$. Otherwise the mask with the highest $q_j$ is selected. Branch S changes both rules (Sec.~\ref{sec:branches}). The branch output $b^k_t$ is the tight axis-aligned box around the selected mask, using the pixel convention of Appendix~\ref{app:readout}. An empty mask gives $b^k_t=\varnothing$.

\subsection{Six memory branches}
\label{sec:branches}

\paragraph{A common read rule.}
D, N, M and R all read memory with DAM4SAM's rule $\rho(t;\mathcal{C},\mathcal{N},r)$, which takes a set $\mathcal{C}$ of conditioning entries, a set $\mathcal{N}$ of ordinary entries, and a grid stride $r$. It first selects up to four conditioning entries: the anchor, the latest entry in $\mathcal{C}$ before frame $t-1$, and then the entries closest to $t$. It fills the remaining slots, up to seven in total, with the latest nonempty entry before $t$, followed by nonempty entries at multiples of $r$, searching backward. Empty, missing and already-selected frames are skipped. Object pointers come from the selected conditioning entries and from the nonempty entries among the 15 preceding frames. A frame is \emph{nonempty} for a branch if that branch predicted at least one foreground pixel there. Table~\ref{tab:branches} defines each branch with this rule.

\begin{table*}[t]
\centering\small
\caption{Branch definitions. $\mathcal{C}_k$ and $\mathcal{N}_k$ are branch $k$'s conditioning and ordinary entries, and $\mathcal{G}_t$ is the set of frames before $t$ on which D and S agreed (both nonempty, box IoU $\ge 0.5$). D, S, M, N and R supply candidate boxes. A supplies arbiter cues only. Frame indices refer to the original video.}
\label{tab:branches}
\begin{tabular}{@{}l >{\raggedright\arraybackslash}p{0.30\textwidth} >{\raggedright\arraybackslash}p{0.27\textwidth} >{\raggedright\arraybackslash}p{0.29\textwidth}@{}}
\toprule
Branch & Memory read for frame $t$ & Update after frame $t$ & Failure it targets\\
\midrule
\textbf{D} & $\rho(t;\mathcal{C}_{\mathrm{D}},\mathcal{N}_{\mathrm{D}},5)$ with $\mathcal{C}_{\mathrm{D}}=\{0\}\cup$ promoted frames (unmodified DAM4SAM). & Store its entry and promote it to $\mathcal{C}_{\mathrm{D}}$ when the test below holds. & Default tracker, robust to nearby distractors.\\
\addlinespace
\textbf{N} & $\rho(t;\{0\},\mathcal{N}_{\mathrm{N}},5)$. & Store its entry and never promote. & A wrong promotion in D, which D never forgets.\\
\addlinespace
\textbf{M} & $\rho(t;\{0\},\mathcal{N}_{\mathrm{M}},1)$, i.e.\ the six latest nonempty frames. & Store its entry. & Rapid appearance change that the five-frame grid misses.\\
\addlinespace
\textbf{S} & Anchor and S's six latest entries that pass SAMURAI's admission test, plus frame $t-1$. & Store its entry, record admission, and update the motion filter. & Jumps to a look-alike away from the predicted motion, and unreliable entries in memory.\\
\addlinespace
\textbf{R} & $\rho\bigl(t;\{0\}\cup(\mathcal{C}_{\mathrm{D}}\cap\mathcal{G}_t),\,\mathcal{N}_{\mathrm{D}}\cap\mathcal{G}_t,\,5\bigr)$, reading D's entries. & Record D--S agreement for frame $t$ and store nothing. & Contaminated entries in D's memory.\\
\addlinespace
\textbf{A} & Anchor only, on frames $t=5,10,15,\ldots$ & None. & Gives the arbiter a drift-free reference to the initial target.\\
\bottomrule
\end{tabular}
\end{table*}

The branches form controlled contrasts. D and N differ only in promotion, so they coincide until D's first promotion. N and M differ only in temporal spacing. R differs from D only by discarding entries that S did not confirm. No branch writes into another branch's memory, and D's state is never modified, so D reproduces standalone DAM4SAM.

\paragraph{Distractor promotion (D).}
D~\cite{dam4sam} promotes frame $t$ to $\mathcal{C}_{\mathrm{D}}$ when four reliability conditions hold: the selected mask is nonempty, its predicted IoU exceeds $0.8$, its area is within 20\% of the median of up to ten nonempty masks among the latest 300 observations, and more than five frames have passed since the previous promotion. It must also pass a distractor test. For each alternative mask, D removes the pixels covered by the selected mask, keeps the largest remaining connected component, and joins it to the selected mask. If the box of this union has IoU at most $0.7$ with the selected box for some alternative, the decoder has found a second plausible object, and D promotes the \emph{selected} target as a distractor-resolving reference.

\paragraph{Motion-guided selection (S).}
S implements SAMURAI~\cite{samurai} in the shared runtime. A constant-velocity Kalman filter on S's own boxes forecasts the box $\widehat B_t$. S treats the target as present when $o>-1$. During initialization and until the filter has 15 consecutive stable frames (selected $q_j>0.3$), S selects the mask with the highest $q_j$. Afterwards it selects
\[
  j^\star=\arg\max_j\;0.85\,q_j+0.15\,\operatorname{IoU}(B_j,\widehat B_t),
\]
where $B_j$ is the box of mask $j$. A selection with $q_{j^\star}<0.3$ resets the stable count. A stored entry is admitted for later reads if $q_{j^\star}>0.5$, $o>0$ and, when a motion score is available, its box overlaps the forecast. S reads the six latest admitted entries, always including frame $t-1$. Its object pointers come from the 15 preceding frames whether or not they were admitted.

\paragraph{Agreement and anchor references (R, A).}
R uses S as an independent check on D. A past entry of D is readable by R only if D and S produced nonempty boxes with IoU at least $0.5$ on that frame, so R requires S to run. R decodes a new mask for frame $t$ from this filtered history and stores nothing, so its errors do not propagate. A decodes from the anchor alone. The arbiter uses A's most recent box and the number of frames since it was produced.

\subsection{Grouping candidates by agreement}
\label{sec:grouping}

On each frame $t>0$ we build a graph on the five candidate branches, $\mathcal{K}=\{\mathrm{D},\mathrm{S},\mathrm{M},\mathrm{N},\mathrm{R}\}$. Branches $k$ and $l$ are joined if both boxes are nonempty with $\operatorname{IoU}(b^k_t,b^l_t)\ge 0.5$, or if both are empty. The connected components are the clusters, so two boxes in one cluster need not overlap by 0.5 themselves; this happens on 0.94\% of the frames of 583 training sequences. %
 Each cluster $C$ has a fixed representative $r(C)$, its first member in the priority order D, S, M, N, R, and offers the box $b^{r(C)}_t$, possibly $\varnothing$. Let $C_{\mathrm{D}}$ be the cluster containing D, whose representative is D. If all five branches form one cluster, the tracker reports $b^{\mathrm{D}}_t$. Otherwise frame $t$ is a \emph{disagreement frame} and the arbiter is run. Because representatives are fixed before scoring, the arbiter chooses among spatial hypotheses rather than among branches.

\subsection{Arbiter}
\label{sec:arbiter}

\paragraph{Cues.}
For each cluster $C$ on a disagreement frame, we compute a cue vector $x_t(C)\in\mathbb{R}^{95}$ from observations up to frame $t$ (Appendix~\ref{app:cues}). Pairwise cues compare $r(C)$ with an opponent. The opponent is D for $C\neq C_{\mathrm{D}}$. For $C_{\mathrm{D}}$ it is the branch that has been outside $C_{\mathrm{D}}$ for the most consecutive frames up to $t$, the first in the order D, S, M, N, R on ties; this count is the disagreement duration used below. %
\begin{itemize}
\item \textbf{Geometry and disagreement (36):} cluster size and membership, disagreement duration, trajectory smoothness, changes in box size and aspect ratio, and mask fragmentation.
\item \textbf{Model confidence (23):} object-presence score, predicted mask IoUs and their margins, mask-logit statistics, changes from recent confidence levels, S's motion state, and the size of $\mathcal{C}_{\mathrm{D}}$.
\item \textbf{Opposing evidence (15):} current and historical overlap between the two masks, how closely one branch's selected mask matches an alternative rejected by the other, and the logits each branch assigns to pixels selected by its opponent.
\item \textbf{Appearance (21):} similarity of mask-pooled image features and object pointers to early target observations, to a template updated on single-cluster frames, and to the representative's last appearance before the disagreement, plus cues from the prompted object pointer, memory-conditioned features, and A's most recent box.
\end{itemize}

\paragraph{Training target.}
Let $u_{t,C}$ be the IoU between the box of $r(C)$ in that branch's own box stream and the ground-truth box. In that stream, a frame without a box carries the branch's own last box forward; the tracker's output instead carries forward the last box it reported (Appendix~\ref{app:empty}). The oracle of Sec.~1 uses the same branch streams. %
The regression target is the frame's contribution to success AUC,
\[
  y_{t,C}=\frac{1}{21}\sum_{k=0}^{20}\bigl[u_{t,C}>k/20\bigr],
\]
where $[\cdot]$ is the Iverson bracket. A regressor $s(\cdot)$ is fitted to predict $y_{t,C}$ from $x_t(C)$ by minimising the squared error
\[
  \textstyle\sum_{(t,C)}\bigl(s(x_t(C))-y_{t,C}\bigr)^2
\]
over the training rows $(t,C)$ described below. The frame average of $y_{t,C}$ is exactly the success AUC, so the regressor is trained for the evaluation measure.

\paragraph{Data and fitting.}
We use one row per cluster on disagreement frames with an even frame index on which the target is present. %
The 350 LaSOT training sequences (933{,}422 frames) yield 151{,}158 rows. The regressor is a histogram gradient-boosted tree model~\cite{sklearn} with 300 iterations, learning rate $0.06$, at most 31 leaves, at least 200 samples per leaf, and $L_2$ regularization $1.0$. These settings come from an earlier geometry-only feasibility study on archived LaSOT-ext, UAV123 and NFS outputs (Appendix~\ref{app:devuse}). Fitting takes about 10\,s on a CPU. Generating the records takes an estimated 37 GPU-hours at 0.141\,s per frame on an L40S, including record writing (Table~\ref{tab:speed}). %

\paragraph{Model selection.}
We chose the cue set and switching rule by ten-fold cross-validation on the training sequences, with folds grouped by object class. The measure was the mean per-sequence gain in success AUC over D on held-out folds. For each model we took its best switching rule, and the 95-cue regressor was kept only if its gain exceeded that of a 12-cue linear control by more than one paired standard error. %
We compared per-frame margins $\delta\in\{0,0.05,0.1,0.2\}$ with rules allowing at most one switch per disagreement episode. All 95 cues with a per-frame margin $\delta=0.05$ performed best. The regressor was then refitted on all 350 sequences and frozen.

\paragraph{Decision rule.}
On a disagreement frame, let $C^\star=\arg\max_C s(x_t(C))$, with ties going to the cluster whose representative comes first in the order D, S, M, N, R. %
The tracker reports
\[
  \hat B_t=\begin{cases}
    b^{r(C^\star)}_t & \text{if } s(x_t(C^\star))-s(x_t(C_{\mathrm{D}}))>\delta,\\
    b^{\mathrm{D}}_t & \text{otherwise.}
  \end{cases}
\]
The decision affects only the output for frame $t$. On frame $t+1$ every branch reads the history prescribed by its own rule. An empty representative produces an empty output, for which the main evaluation protocol carries the previous reported box forward (Appendix~\ref{app:empty}). A tie leaves the output at DAM4SAM's prediction.

\paragraph{Causality.}
All cues use only frames up to $t$. Four appearance cues use a template from D's first five tracking frames, and on frames 1--4 they use the frames observed so far. A causal replay matched the online decisions on all 15{,}897 frames of five training runs and matched the earlier whole-record decisions on every LaSOT evaluation frame. %

\paragraph{Cost.}
Beyond the shared encoder pass, each of N, M and S adds one memory attention, mask decoding and memory encoding per frame. R adds memory attention and mask decoding per frame, and A adds them every fifth frame. Across four online training runs, the CPU arbiter took 0.001--0.007\,s per frame averaged over all frames and 0.02--0.32\,s averaged over disagreement frames. The slowest single call in each run took 0.03--0.43\,s. The arbiter's cost grows with sequence length; the slowest run had 11{,}397 frames. The branch timing of Section~\ref{sec:speed} excludes the arbiter, and Appendix~\ref{app:sentry} gives the output rule of PARMA (SENTRY-D4S-led), which reports SENTRY-D4S by default. %

\section{Experimental Setup}
\label{sec:setup}

\paragraph{Benchmarks and baselines.} We evaluate on LaSOT~\cite{lasot} (280 test sequences), LaSOT-ext~\cite{lasotext} (150 sequences of 15 classes absent from LaSOT), UAV123~\cite{uav123} (123), NFS~\cite{nfs} (100, 30\,fps) and TNL2K~\cite{tnl2k} (699 of the 700 test sequences, first-frame box only; Appendix~\ref{app:branchdetails}). We re-ran SAM\,2.1~\cite{sam2}, SAMURAI~\cite{samurai}, HiM2SAM~\cite{him2sam}, DAM4SAM~\cite{dam4sam} and SENTRY-D4S~\cite{sentry} (default configuration of its public release) from their official code and archived their masks; SAM2Long has no box-tracking code. On TNL2K we do not compare with the baselines. \emph{D (same run)} is the D branch of our own run. Our output equals its box whenever D is chosen, so it is the paired reference for what the arbiter adds.

\paragraph{Read-out and scoring.} Each system's box is the tight rectangle around its mask (read-out \ReadU, width $x_{\max}-x_{\min}+1$), and frame 0 is the given box. All systems are scored with one port of the official LaSOT protocol: a missing box is replaced by the previous one; as in the official code, frames whose ground-truth box has a non-positive coordinate count as failures for success AUC but as hits for the two precision measures, and frames marked absent count as failures for every system. We report success AUC, normalised precision AUC ($P_\text{norm}$) and precision at 20 pixels ($P$), all $\times 100$. The NFS baselines ran under a different torch version, so absolute NFS scores are not comparable with the literature. %

\paragraph{Statistics and registration.} Comparisons are paired per sequence and written as estimate (s.e.); better/worse counts sequences that change by more than 2 AUC points. For each test, we fixed the readings, the pass lines and the allowed wording in a time-stamped file before any of its results existed (Appendix~\ref{app:registration}); registered tests use a Holm correction~\cite{holm}. An online run of the complete tracker on all 280 LaSOT test sequences reproduced the offline decisions on every frame (Appendix~\ref{app:online}). LaSOT results are descriptive, because the main configuration was chosen after a one-time evaluation on LaSOT; Appendix~\ref{app:devuse} lists how each test set had been used before its test.

\section{Results}

\begin{figure}[t]
\centering
\includegraphics[width=0.92\columnwidth]{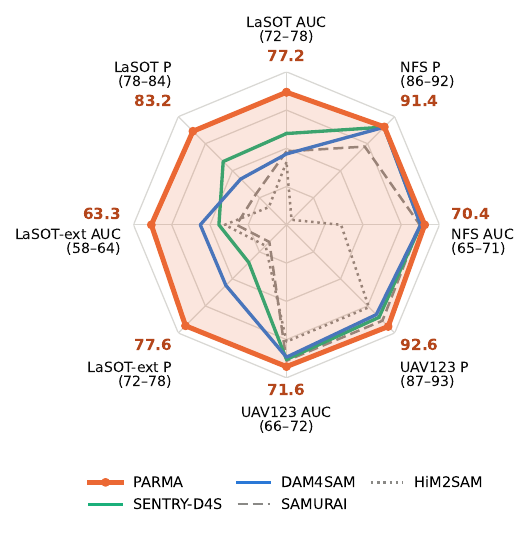}
\caption{PARMA and the four strongest re-run SAM\,2.1-L trackers (Tables~\ref{tab:main} and~\ref{tab:mainfull}) on success AUC and precision $P$ of four benchmarks; SAM\,2.1, far below the others, is omitted. Every axis spans 6 points (range in parentheses); PARMA's values in orange.}
\label{fig:radar}
\end{figure}

\begin{table}[t]
\centering
\caption{Success AUC ($\times 100$) under one read-out (\ReadU) and the official protocol. Deltas are relative to SAM\,2.1; \textbf{bold}: best. The five baselines are separate re-runs from official code. $^\dagger$Registered test passed (Appendix~\ref{app:confirm}); LaSOT is descriptive. $^\ast$Absolute NFS scores are not comparable with the literature. Table~\ref{tab:mainfull} adds $P_\text{norm}$, $P$, the D branch of the same run and reported results from other papers.}
\label{tab:main}
{\footnotesize\setlength{\tabcolsep}{3pt}%
\begin{tabular}{@{}lcccc@{}}
\toprule
Method & LaSOT & LaSOT-ext & UAV123 & NFS$^\ast$\\
\midrule
SAM\,2.1~\cite{sam2} & 69.49 & 57.24 & 68.84 & 69.46\\ %
\midrule
SAMURAI~\cite{samurai} & 74.88\,\gain{+5.39} & 59.96\,\gain{+2.72} & 71.33\,\gain{+2.49} & 70.23\,\gain{+0.77}\\
HiM2SAM~\cite{him2sam} & 74.46\,\gain{+4.97} & 60.42\,\gain{+3.18} & 70.58\,\gain{+1.74} & 67.14\,\loss{$-$2.32}\\
DAM4SAM~\cite{dam4sam} & 74.78\,\gain{+5.29} & 61.38\,\gain{+4.14} & 71.19\,\gain{+2.35} & 70.26\,\gain{+0.80}\\
SENTRY-D4S~\cite{sentry} & 75.58\,\gain{+6.09} & 60.65\,\gain{+3.41} & 71.28\,\gain{+2.44} & 70.27\,\gain{+0.81}\\
\midrule
PARMA & \textbf{77.20}\,\gain{+7.71} & \textbf{63.30}$^\dagger$\,\gain{+6.06} & \textbf{71.56}\,\gain{+2.72} & \textbf{70.42}\,\gain{+0.96}\\
\bottomrule
\end{tabular}}
\end{table}

\subsection{Main results}
\label{sec:main}
PARMA sets a new state of the art among SAM\,2-based trackers on LaSOT (77.2 AUC) and LaSOT-ext (63.3), above all five re-run baselines and every SAM\,2-based result we found (lower block of Table~\ref{tab:mainfull}); under the more permissive SAMURAI-style valid-frame rule, its LaSOT-ext AUC is 64.88 against the highest such result, 62.97~\cite{samosa}. It scores highest on every benchmark in AUC, $P_\text{norm}$ and $P$ (Tables~\ref{tab:main} and~\ref{tab:mainfull}). On LaSOT it gains +2.38\se{0.45} AUC over D (same run) and +1.62\se{0.39} over SENTRY-D4S, the strongest re-run baseline (+1.27\se{0.38} on the 235 sequences outside the subset V used during development). On LaSOT-ext it gains +2.65\se{0.66} over D (same run) and +1.92\se{0.71} over DAM4SAM. %
The registered test, against D (same run) and SENTRY-D4S, holds pooled over LaSOT-ext, UAV123 and NFS (+1.06\se{0.23} and +1.02\se{0.23}) and on LaSOT-ext alone. On UAV123 and NFS PARMA improves on both by 0.15--0.28 AUC, which is not established on either alone (Appendix~\ref{app:confirm}); there the branches disagree on only about 4\% of frames (Sec.~\ref{sec:why}). %
Training on all 1{,}120 LaSOT training sequences gives nearly the same results (Appendix~\ref{app:ablation}). SAM\,3-based trackers, built on a newer foundation model, report higher numbers~\cite{sam3rl,rethinkmem}; we will report PARMA on SAM\,3 in a later version.

\subsection{External test on TNL2K}
\label{sec:tnl2k}
TNL2K played no part in the design. Before any tracking job on it, we registered two readings: (i) the gain of the arbiter over its own base tracker, and (ii) its gain over an untrained consensus rule that switches about as often. The consensus rule is the \emph{medoid}, which reports the branch whose box agrees most with the other four; its conservative version reports it only if its agreement exceeds D's by more than $\tau$. We fixed $\tau = 0.4$ on 583 training sequences, where the rule leaves D's cluster on 6.51\% of frames and the arbiter on 6.55\%. On TNL2K the two rates are 6.64\% and 8.97\% (Appendix~\ref{app:tnl2k}). %

On TNL2K, PARMA gained 1.89 AUC points (s.e.\ 0.26) over its own DAM4SAM branch and 1.32 (s.e.\ 0.21) over the calibrated consensus rule. Both comparisons met their pre-registered thresholds (Table~\ref{tab:tnl2k}). The consensus rule itself gains only +0.57 and makes 57 sequences worse. Choosing the best cluster with ground truth would gain +3.30; the arbiter obtains 57\% of it. On TNL2K, PARMA reaches 62.73 AUC, above the highest SAM\,2-based result we found (61.7, SAMITE-L as measured in~\cite{rethinkmem}). %

\subsection{Learned versus untrained arbitration}
\label{sec:learned}
To measure how much of the gain needs learning, Table~\ref{tab:delta} compares the arbiter with the conservative medoid at equal switching rates on 583 held-out training sequences. Across these sequences, the plain medoid and the learned arbiter gained 2.39 and 2.36 AUC points over DAM4SAM, respectively. AUC fell by more than two points on 56 sequences with the plain medoid and on 16 with the arbiter. At each margin in Table~\ref{tab:delta}, the arbiter gains more than the consensus rule switching equally often and makes fewer sequences worse. On the development benchmarks, learning adds +0.87\se{0.35} over the plain medoid on LaSOT and +0.38\se{0.19} pooled over LaSOT-ext, UAV123 and NFS (Appendix~\ref{app:ablation}). Much of the gain thus comes from running diverse memory policies; what learning adds above all is avoiding switches that badly hurt a video. %

\begin{table}[t]
\centering
\caption{Learned arbiter vs consensus at equal switching rate (583 held-out training sequences). Switched: frames whose box lies outside D's cluster. Gain over D (s.e.) and sequences worse by more than 2 points. The consensus rule is interpolated along $\tau$ to the arbiter's switching rate, so its counts are not integers.}
\label{tab:delta}
\resizebox{\columnwidth}{!}{%
\begin{tabular}{@{}lcccc@{}}
\toprule
 & & \multicolumn{2}{c}{Arbiter} & Consensus\\
\cmidrule(lr){3-4}\cmidrule(l){5-5}
$\delta$ & Switched & Gain & Worse & Gain / worse\\
\midrule
0.05 (PARMA) & 6.55\% & \up{+2.36}\se{0.33} & 16 & \up{+2.20} / 39.4\\ %
0.1 & 5.83\% & \up{+2.29}\se{0.32} & 11 & \up{+1.98} / 26.9\\
0.2 & 4.32\% & \up{+1.98}\se{0.30} & 7 & \up{+1.72} / 16.6\\
0.3 & 3.04\% & \up{+1.61}\se{0.27} & 6 & \up{+1.44} / 9.0\\
\midrule
\multicolumn{2}{@{}l}{Plain medoid (7.25\%)} & \multicolumn{3}{c}{\up{+2.39}\se{0.39}, 56 worse}\\
\bottomrule
\end{tabular}}
\end{table}

\subsection{Ablations, cost and memory}
\label{sec:ablation}

\begin{table*}[t]
\centering
\caption{Branch subsets (all with anchor A). Time: per-frame time relative to D alone on L40S, derived from per-branch timings, excluding record writing. Train: gain over D on 583 held-out training sequences (share of the full gain). Test: gain over D (same run). Pooled: LaSOT-ext, UAV123 and NFS.}
\label{tab:branchabl}
\resizebox{\textwidth}{!}{%
\begin{tabular}{@{}ccccc cc ccccc@{}}
\toprule
D & S & M & N & R & Time & Train (583) & LaSOT & LaSOT-ext & UAV123 & NFS & Pooled\\
\midrule
\cm & & \cm & & & 1.21$\times$ & \up{+2.07}\se{0.31}, 88\% & \up{+1.85}\se{0.39} & \up{+1.61}\se{0.59} & \up{+0.03}\se{0.07} & \up{+0.11}\se{0.06} & \up{+0.58}\se{0.20}\\ %
\cm & & \cm & \cm & & 1.39$\times$ & \up{+2.18}\se{0.32}, 92\% & \up{+2.02}\se{0.44} & \up{+2.18}\se{0.60} & \up{+0.05}\se{0.08} & \up{+0.17}\se{0.11} & \up{+0.80}\se{0.20}\\
\cm & \cm & \cm & & \cm & 1.59$\times$ & \up{+2.26}\se{0.33}, 96\% & \up{+2.31}\se{0.44} & \up{+2.28}\se{0.63} & \up{+0.27}\se{0.12} & \up{+0.22}\se{0.14} & \up{+0.92}\se{0.22}\\
\cm & \cm & \cm & \cm & \cm & 1.77$\times$ & \up{+2.36}\se{0.33}, 100\% & \up{+2.38}\se{0.45} & \up{+2.65}\se{0.66} & \up{+0.28}\se{0.15} & \up{+0.24}\se{0.14} & \up{+1.06}\se{0.23}\\
\bottomrule
\end{tabular}}
\end{table*}

\begin{table}[t]
\centering
\caption{Cue groups (training sequences, 583; nested; full branch set).}
\label{tab:cues}
\resizebox{\columnwidth}{!}{%
\begin{tabular}{@{}lcccc@{}}
\toprule
Cues & \# & Gain over D & Share & vs full\\
\midrule
Geometry \& disagreement & 36 & \up{+2.05}\se{0.31} & 87\% & \down{$-$0.31}\se{0.09}\\ %
+ model confidence & 59 & \up{+1.98}\se{0.30} & 84\% & \down{$-$0.38}\se{0.10}\\
+ opposing evidence & 74 & \up{+1.97}\se{0.30} & 84\% & \down{$-$0.39}\se{0.09}\\
+ appearance (all) & 95 & \up{+2.36}\se{0.33} & 100\% & 0\\
12-cue linear control & 12 & \up{+1.87}\se{0.30} & 79\% & \down{$-$0.49}\se{0.11}\\
\bottomrule
\end{tabular}}
\end{table}

\paragraph{Branches.} Configurations with a larger compute budget gain more on average (Table~\ref{tab:branchabl}); the rows are not nested one branch at a time, since D+S+M+R drops N and adds S and R. On LaSOT most of the gain is available from two candidates, D and M, plus the read-only anchor A run every fifth frame: +1.85\se{0.39}, 78\% of the full gain, at $1.21\times$ the time of D. On LaSOT-ext the full set is clearly ahead (D+M trails it by 1.04\se{0.45}), and on UAV123 D+M gains almost nothing. %

\paragraph{Cues.} On the held-out training sequences, geometry and disagreement cues alone gained 2.05 AUC points over DAM4SAM (Table~\ref{tab:cues}). The gain was 1.97 after adding model-confidence and opposing-evidence cues, and 2.36 with the appearance cues, which include the object pointer. A 12-cue linear control reaches +1.87, and with only 90 training sequences the gain is already +2.21 to +2.32 (Appendix~\ref{app:ablation}). %

\paragraph{Cost and memory.} Measured within the same job, the full system takes 1.89--2.08$\times$ the per-frame time of D alone on L40S (four runs), i.e.\ about $2\times$. The jobs that produced our records ran at 0.141\,s per frame on L40S, including record writing but not the arbiter; SENTRY-D4S in its public default configuration ran at 0.510\,s in separate runs (Appendix~\ref{app:ablation}). Peak GPU memory is within about 1\% of DAM4SAM alone (Appendix~\ref{app:ablation}). %

\begin{figure}[t]
\centering
\includegraphics[width=\columnwidth]{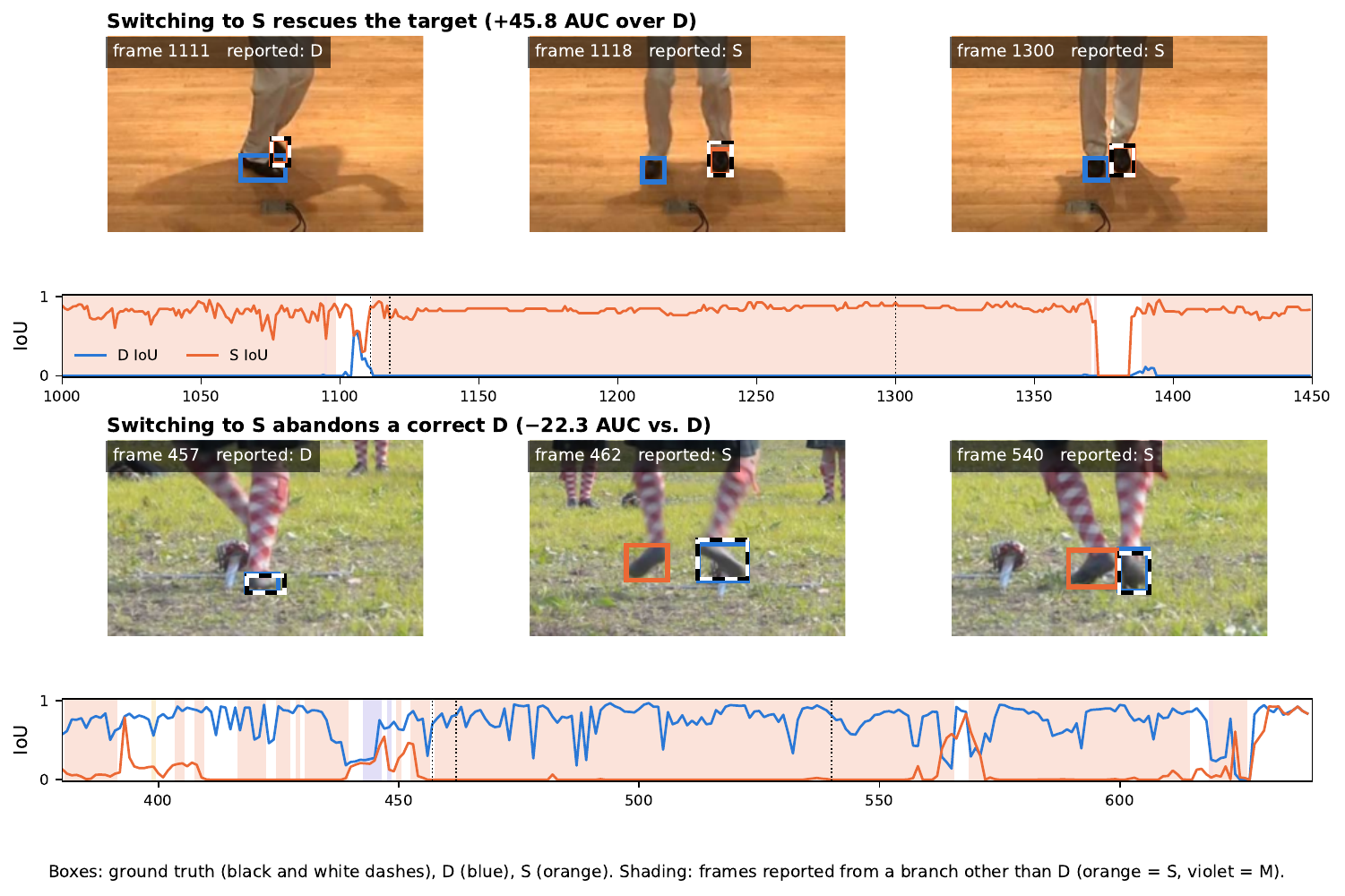}
\caption{A matched success and failure on LaSOT-ext. Top, dancingshoe-8 (+45.8 AUC over D): D has lost the target by frame 1000, and the arbiter reports S, which stays on it. Bottom, dancingshoe-4 ($-$22.3): from frame 458 three branches, S among them, move together to an identical second shoe while D alone stays on the target, and the arbiter follows the majority. Below each row: IoU of D and S with the ground truth.}
\label{fig:dancingshoe}
\end{figure}

\subsection{Arbiter mechanism}
\label{sec:why}
Averaged over sequences, the output differs from D's on 5.90\% of frames on LaSOT, 12.74\% on LaSOT-ext, 2.36\% on UAV123, 1.03\% on NFS and 8.97\% on TNL2K; pooled over all frames of the 653 test sequences of the first four, the arbiter reports a branch other than D on 8.0\% of frames. %
 Frames reported from S account for 79\% of the gain over D (74\% on LaSOT, 82\% on LaSOT-ext); M, N and R contribute the rest. The gain is concentrated in long switches: runs of 31 frames or more are 2.5\% of all switched runs but bring 61\% of the gain, while single-frame switches bring 6\%.

On 583 held-out training sequences, a depth-3 decision tree fitted to the arbiter's decisions retains 96\% of its gain and reads as a short rule. A cluster of three or more branches is chosen unless its mask fills little of its box without a clearly better appearance match, or it agrees with anchor A much less than D does. A cluster of one or two branches is chosen only when its object pointer is closer to the prompt's than D's is and it agrees better with A. Shuffling the appearance cues removes 0.99 of the 2.36-point gain, the geometry cues 0.67. On TNL2K, the gain is largest on sequences with out-of-view (+4.03), motion blur (+3.76) and aspect-ratio change (+3.14), and smallest under low resolution (+0.71) (Appendix~\ref{app:interp}). Appendix~\ref{app:examples} lists the largest gains and losses. Fig.~\ref{fig:dancingshoe} shows a matched pair. On dancingshoe-8, the arbiter follows S after D loses the target. On dancingshoe-4, three branches switch to the other, identical shoe and the arbiter follows them, while D stays on the target. %

\section{Limitations}
The arbiter only chooses, so it gains little where all branches fail together or rarely disagree, as on UAV123 and NFS, and because it never writes back it cannot help D recover. LaSOT results are descriptive, and the other development benchmarks had been used before their registered test (Appendix~\ref{app:devuse}); TNL2K is the only untouched benchmark, and there we do not compare with the baselines. Four of our five LaSOT re-runs are 0.32--0.72 AUC below their published numbers for reasons we could not fully explain (Appendix~\ref{app:readout}). Appendix~\ref{app:limits} discusses further limitations.

\section{Conclusion}
Different SAM\,2 memory rules fail on different frames. Running several of them side by side on one frozen model and learning only which output to report turns this complementarity into accuracy, at about twice the per-frame cost of one tracker. The learned arbiter improves its own DAM4SAM branch on every benchmark we tested, most clearly on LaSOT-ext and on the external TNL2K test; on UAV123 and NFS, where the branches rarely disagree, the gains are small. Compared with untrained consensus, it mainly avoids switches that badly hurt a video.

\paragraph{Acknowledgements.} This work began as a course project in CS\,566 at the University of Wisconsin--Madison. This research used the computing resources and services of the Center for High Throughput Computing (CHTC) at the University of Wisconsin--Madison~\cite{chtc}.
SAM\,2 and the code of SAMURAI, HiM2SAM, DAM4SAM and SENTRY are released under the Apache License 2.0, and we used them under its terms. We used the LaSOT, LaSOT-ext, UAV123, NFS and TNL2K benchmarks for research evaluation only.

\clearpage
\appendix
\onecolumn

\section{Pre-registration and what each test decided}
\label{app:registration}

All registration files are time-stamped and were written before the results they govern existed. Frozen artifacts are identified by SHA-256: the arbiter file (2c83ffb6\ldots; for \#12, the SAM\,3 arbiter file d30e7fb8\ldots), the scorer files and the evaluation scripts. Every evaluation script verifies these hashes and stops on a mismatch. Each registered reading was evaluated once on the test sequences, and its outputs are archived unchanged. Except for TNL2K, the test sequences were not untouched before these evaluations; see ``Development use of the test sets'' below.

\begin{table}[h]
\centering\small
\caption{Registered evaluations. Hash prefixes are those of the registration files (\#1 is identified by its date).}
\label{tab:registration}
\begin{tabular}{@{}c p{0.17\textwidth} p{0.2\textwidth} p{0.29\textwidth} p{0.22\textwidth}@{}}
\toprule
\# & What & Registration & Primary reading and rule & Outcome\\
\midrule
1 & LaSOT 280, one-time evaluation & 2026-11-21 00:48, before the final records were delivered & SENTRY-led $-$ SENTRY-D4S (separate run). ``Holds'' if $>0$ and $>2$ s.e.\ on 280 \emph{and} $>0$ on the 235 outside V & \textbf{Holds}: +1.21\se{0.32}, 3.8 s.e.; 235: +0.89\se{0.28}\\ %
2 & Main method chosen & after \#1 & DAM4SAM-led adopted because it scored higher in \#1 (0.7717 vs 0.7679) & not a test; LaSOT results of the main method are descriptive\\
3 & LaSOT-ext, UAV123, NFS & 3b9d3f87\ldots, 2026-11-21 16:46, before any tracking job of ours on them & (i) and (ii) must both exceed their Holm-corrected multiple; pooled readings $>2$ s.e., no benchmark below 0 by $>2$ s.e. & Pooled \textbf{holds}; LaSOT-ext \textbf{exceeds}; UAV123, NFS \textbf{positive, not established}\\ %
4 & LaSOT strictly causal recomputation & dbbf5f59\ldots & Explanatory; same records as \#1 & Choice identical on every frame; 0.7720 (main empty-frame convention), 0.7717 (earlier one)\\ %
5 & Supplementary readings & e13dfa6b\ldots & Explanatory & Table~\ref{tab:mainfull}\\
6 & Test-set ablation & 975c1cdc\ldots & All rows reported; writing rules for the medoid fixed in advance & Tables~\ref{tab:branchabl},~\ref{tab:rules}\\ %
7 & Read-out U+1 & e1123784\ldots & Explanatory; verdicts stay in \ReadU & Table~\ref{tab:uplus1}\\
8 & LaSOT-ext, relaxed valid-frame rule & 1874f8a4\ldots, 2026-11-22; baselines under this rule already seen, our outputs not & Explanatory, no pass line; verdicts stay under the official rule & Table~\ref{tab:relaxed}\\
9 & Online LaSOT 280, side by side with SENTRY-D4S & 2026-11-21 14:53, before the online jobs were scored & SENTRY-led $-$ SENTRY-D4S of the same job; ``holds'' if $>0$ and $>2$ s.e.\ on 280 \emph{and} $>0$ on the 235 outside V & \textbf{Holds}: +1.70\se{0.37}, 4.57 s.e.; 235: +1.30\se{0.35} (Table~\ref{tab:online})\\ %
10 & TNL2K, external test (699 of 700) & 2ea2ea7d\ldots, 2026-11-22 23:49, before any tracking job on TNL2K & (i) ours $-$ D (same run); (ii) ours $-$ conservative medoid, $\tau = 0.4$. Holm: 2.24 and 2.0 s.e.; a failed sequence voids both & \textbf{Both hold}: (i) +1.89\se{0.26}, 7.20 s.e.; (ii) +1.32\se{0.21}, 6.40 s.e.\ (Table~\ref{tab:tnl2k})\\ %
11 & Arbiter refitted with seeds 1--4 & 1d9609de\ldots, 2026-11-24 09:30, before any seed other than 0 was fitted & Explanatory; only the regressor's random seed changes; seed 0 must reproduce the archived boxes byte for byte & Reproduced; s.d.\ of mean AUC at most 0.05 on every SAM\,2.1 benchmark (Appendix~\ref{app:ablation})\\ %
12 & SAM\,3: LaSOT 280, LaSOT-ext 150 & aa9b6731\ldots, 2026-11-22 13:30, before any SAM\,3 test record existed and any selector was trained on SAM\,3; supplements 4cf1ef96\ldots{} (before any test record) and 37b936bd\ldots{} (stricter; while test records were being produced, none opened) & (i) ours $-$ D of the same record; Holm: 2.24 and 2.0 s.e.\ ``Higher than published'' only if (i) holds and the mean AUC exceeds the highest published number & \textbf{Both hold}: +0.78\se{0.31}, 2.50 s.e.; +1.01\se{0.46}, 2.22 s.e.; not higher than published\\ %
\bottomrule
\end{tabular}
\end{table}

Registration \#12 concerns a SAM\,3 version of the tracker; its results will be reported in a later version.

\paragraph{Registration files and scripts.} Table~\ref{tab:regfiles} gives the full SHA-256 of each registration file and of the evaluation script that applied it. The files are written in Chinese; Table~\ref{tab:regfiles} fixes their content by SHA-256, so that it can be checked when they are released. Our evidence for the registration times is the files' time stamps and our time-stamped project logs. The files in the registration folder were also committed to a private Git repository, all but one on 2026-11-23 at 00:46 (UTC$-$6) and the seed-check registration (\#11) on 2026-11-24 at 13:26. For TNL2K (\#10) this commit precedes every tracking job on TNL2K; for the earlier registrations it only shows that the files existed by then. The readout of \#1 is a file in the project folder outside that registration folder, and \#9 is a dated entry in our project log. %

\begin{table}[h]
\centering\scriptsize
\caption{Registration files (full SHA-256) and the evaluation script of each registered evaluation. Earlier versions of \#3 (2bb7259d\ldots, fe22bae4\ldots) and \#4 (ebb1278d\ldots) were replaced by the versions listed.}
\label{tab:regfiles}
\begin{tabular}{@{}cll@{}}
\toprule
\# & File & SHA-256\\
\midrule
1 & LaSOT one-time evaluation, readout (2026-11-21 00:48) & \texttt{41b8a026479ffb431fc14aeb196ee24e592381293c13fc4b938f0fff0c2c63fc}\\
  & script \texttt{final\_eval\_280.py} & \texttt{044eac35ed741c73273f1d2e9f6b997cd50af852ef24f7daf6b22c4f88f58429}\\
3 & LaSOT-ext, UAV123, NFS readout, final version (2026-11-21 16:46) & \texttt{3b9d3f879e2f8164e14487d28e5e337e1463558218986017ecaecf2034dd2df6}\\
  & script \texttt{score\_bench3\_20261121.py} & \texttt{2fbc791d019211aece3503e2ed498552bed35cbba42310c042228df0d87882a3}\\
4 & LaSOT strictly causal recomputation, readout version 2 & \texttt{dbbf5f59940fb8f4cc632973fde5ac209da199b000c4ca06ce8bc5589de6bdea}\\
  & manifest of the input record packages & \texttt{157ab511519784bdd5ed584c911788956117f3b1c598f01c2395bcf33632cec1}\\
  & manifest of the box files of \#1 & \texttt{3f64430f7d01dda85c5a237d9b22a29e0168880cbba1a1f15f63050eda161872}\\
  & script \texttt{score\_lasot\_causal\_20261121.py} & \texttt{6a131f46f61c43f2c57c7822859d9b1fec4d2673d136bef2ab4b7ef9606dd6be}\\
5 & Supplementary readings & \texttt{e13dfa6b6186ad76420c2f3d29a16ef0a6d6ec7927a82b7c6fc1f7fdbeb3914e}\\
  & script \texttt{supplementary\_readings\_20261122.py} & \texttt{34f9a398cd1ac3d687eff2e306c964da18b3cfb7ac2cd6882e7669a4d3d1614c}\\
6 & Test-set ablation & \texttt{975c1cdca6bf84acd40961defd631f5f4d670d690f13a1b046b49ef37c17b3c1}\\
  & script \texttt{ablation\_test\_20261122.py} & \texttt{057b673242d01b410c940088e27109ffbe493a472464faf74f65513a1669e7c5}\\
7 & Read-out U+1 & \texttt{e1123784257955a4ecccb5f6bdafee6d390f75fa50066e741818e14e3df04d08}\\
  & script \texttt{readout\_plus1\_20261122.py} & \texttt{066ebed6c16b3718ba10490e7a13c3996fa45d82fe677012c1ceb5e3b21c6c4b}\\
8 & LaSOT-ext, relaxed valid-frame rule & \texttt{1874f8a40fda1cf85ae08f612f7f2bc868a3826cadb6dbebdfed9bcf0f3a1de5}\\
  & script \texttt{lasotext\_relaxed\_rule\_20261122.py} & \texttt{cc190252ccdb068ba8e7a61cab11665ca003c99046080cebf5ada337905ffc75}\\
9 & Online LaSOT run: readout in the project log (2026-11-21 14:53) & --\\
  & script \texttt{score\_online280\_20261121.py} & \texttt{00f6cf5e6183c0fc8b65d01a9d1e879628adcc584e73c700ac16e9426aacb877}\\
10 & TNL2K & \texttt{2ea2ea7d2c9bb163b61354f8657048389e4fd5405925081e7a663944518ce52d}\\
  & script \texttt{score\_tnl2k\_20261123.py} & \texttt{795422cacd994d761e8eb4a194781e111c74eec08760809a3b181a60ad55a092}\\
11 & Seed check & \texttt{1d9609de3348af8e774510421b07d3a26688a5de4acac3e4f03518e1573d7f81}\\
  & script \texttt{seed\_check\_20261124.py} & \texttt{8c79f6d1c71724dbff4123c881acafac1179a55023f2d7329a7cafc0f008b28c}\\
12 & SAM\,3 & \texttt{aa9b67317fe692e9623e653c0907b860bdd4891959e33aa5f889752c27f18eaa}\\
  & supplement 1 & \texttt{4cf1ef968e035de49dc14c8b5dfc87657857d859132fc5dbc4b1ec313675403d}\\
  & supplement 2 & \texttt{37b936bd850cb34cb862167293f352de52cb017a93ae1729402b542104e84b2c}\\
  & script \texttt{score\_sam3\_20261123.py} & \texttt{6fec88a9f27c3c6c2c8c10e1a0f4cb149b35b584bc7439b04fbfc3e46e5c3d6c}\\
\bottomrule
\end{tabular}
\end{table}

\paragraph{Writing rules fixed in advance for the test-set ablation.}
\begin{itemize}
\item If the main method exceeds the medoid by more than 2 s.e.\ on LaSOT or pooled, we may state how much learning adds over the strongest untrained rule. On LaSOT it does: +0.87\se{0.35}, 2.5 s.e. Pooled, it is +0.38\se{0.19}, 2.04 s.e., also just above the line.
\item If the medoid exceeds the main method by more than 2 s.e.\ anywhere, this must be stated in the text. It does not, on any benchmark.
\item The reduced configurations are reported as speed--accuracy trade-offs only.
\end{itemize}

\paragraph{Development use of the test sets.}
\label{app:devuse}
\begin{itemize}
\item \textbf{LaSOT.} V is a subset of 45 LaSOT test sequences, fixed in advance and sampled stratified by difficulty. %
It was evaluated several times while the method was being developed: six occasions are recorded, the last being registered evaluation \#1. Because V was stratified by difficulty and used during development, its numbers (the V row of Table~\ref{tab:subsets}) are skewed towards hard sequences and must not be cited as a benchmark result. PARMA was not evaluated on the other 235 sequences before \#1. Descriptive statistics and ground-truth upper bounds had, however, been computed on the archived baseline outputs of all 280 sequences. A further 13 of the 235 had been looked at before V was drawn, so 222 were never touched. The analysis that motivated the method used LaSOT test sequences.
\item \textbf{LaSOT-ext, UAV123, NFS.} Before \#3 was registered:
  \begin{itemize}
  \item descriptive statistics and upper bounds had been computed on the five archived baselines' outputs;
  \item a geometry-only gradient-boosted selector, a feasibility probe, had been fitted to those outputs and influenced the design;
  \item the decision to add the anchor branch A drew on diagnostics using 36 LaSOT-ext and 12 UAV123 test sequences with ground truth.
  \end{itemize}
  No parameter fitted on these benchmarks entered the frozen arbiter.
\item \textbf{TNL2K.} Not used for any design decision before \#10. The project had never used TNL2K, and when the readings were registered the data were still being downloaded and no sequence had been tracked. The value $\tau = 0.4$ of the conservative medoid was fixed from training sequences only. The scoring script was reviewed independently twice and frozen before any box of a TNL2K record was read or scored; until then only structural checks had been run on the records. It was then run once as a dry run, once as a rehearsal of the full pipeline that reports no numbers, and once for the result.
\item \textbf{SAM\,3 (\#12).} The same LaSOT and LaSOT-ext test sequences as above, so their prior use carries over. On SAM\,3 itself, the test records were only checked for completeness until the arbiter files were frozen, and no box of a test record was read or scored before then. The scoring script was frozen and run in the same three steps as for TNL2K. %
\end{itemize}

\begin{table}[h]
\centering\small
\caption{LaSOT by subset (PARMA, 350 training sequences, strictly causal, main convention).}
\label{tab:subsets}
\begin{tabular}{@{}lcccccc@{}}
\toprule
Subset & $n$ & AUC & $-$ D (same run) & $-$ SENTRY-D4S (separate run) & $-$ DAM4SAM (separate run) & Frames switched\\
\midrule
All & 280 & 77.20 & \up{+2.38}\se{0.45}, 43/6 & \up{+1.62}\se{0.39}, 42/14 & \up{+2.42}\se{0.45}, 48/5 & 5.90\%\\ %
Outside V & 235 & 78.50 & \up{+1.73}\se{0.40}, 28/5 & \up{+1.27}\se{0.38}, 31/11 & \up{+1.83}\se{0.40}, 34/4 & 4.84\%\\
V & 45 & 70.41 & \up{+5.78}\se{1.83}, 15/1 & \up{+3.43}\se{1.45}, 11/3 & \up{+5.54}\se{1.77}, 14/1 & 11.42\%\\
\bottomrule
\end{tabular}
\end{table}

\section{Full results, registered tests and LaSOT in detail}
\label{app:confirm}
Table~\ref{tab:mainfull} gives the full results behind Table~\ref{tab:main}: AUC, $P_\text{norm}$ and $P$ on each benchmark, the D branch of the same run, and reported SAM\,2.1-L results from other papers.

\begin{table*}[t]
\centering
\caption{Full results behind Table~\ref{tab:main}, under one read-out (\ReadU) and the official protocol ($\times 100$). Best in \textbf{bold}, second \second{underlined}; green: gain over D (same run). The five baselines are separate re-runs from official code. $^\dagger$Registered test passed (Table~\ref{tab:confirm}); the pooled test over LaSOT-ext, UAV123 and NFS also holds. LaSOT is descriptive. $^\ast$Absolute NFS scores are not comparable with the literature. The lower block lists reported SAM\,2.1-L results from other runs and scorers (AUC only; source and valid-frame rule in parentheses).}
\label{tab:mainfull}
\resizebox{\textwidth}{!}{%
\begin{tabular}{@{}l ccc ccc ccc ccc@{}}
\toprule
 & \multicolumn{3}{c}{LaSOT} & \multicolumn{3}{c}{LaSOT-ext} & \multicolumn{3}{c}{UAV123} & \multicolumn{3}{c}{NFS$^\ast$}\\
\cmidrule(lr){2-4}\cmidrule(lr){5-7}\cmidrule(lr){8-10}\cmidrule(l){11-13}
Method & AUC & $P_\text{norm}$ & $P$ & AUC & $P_\text{norm}$ & $P$ & AUC & $P_\text{norm}$ & $P$ & AUC & $P_\text{norm}$ & $P$\\
\midrule
SAM\,2.1~\cite{sam2} & 69.49 & 71.88 & 73.56 & 57.24 & 62.05 & 69.30 & 68.84 & 76.70 & 89.18 & 69.46 & 79.42 & 89.74\\ %
SAMURAI~\cite{samurai} & 74.88 & 77.51 & 79.70 & 59.96 & 64.97 & 72.95 & \second{71.33} & \second{79.48} & \second{92.32} & 70.23 & 80.17 & 90.34\\
HiM2SAM~\cite{him2sam} & 74.46 & 76.72 & 78.98 & 60.42 & 65.10 & 73.16 & 70.58 & 78.65 & 91.57 & 67.14 & 76.60 & 86.28\\
DAM4SAM~\cite{dam4sam} & 74.78 & 78.12 & 80.54 & \second{61.38} & \second{66.84} & \second{75.36} & 71.19 & 79.25 & 91.98 & 70.26 & 80.92 & 91.41\\
SENTRY-D4S~\cite{sentry} & \second{75.58} & \second{78.95} & \second{81.51} & 60.65 & 66.04 & 74.09 & 71.28 & 79.40 & 92.13 & \second{70.27} & \second{80.93} & \second{91.42}\\
D (same run) & 74.82 & 78.16 & 80.65 & 60.65 & 66.07 & 74.13 & 71.28 & 79.42 & 92.20 & 70.18 & 80.83 & 91.28\\
\midrule
PARMA & \textbf{77.20}\,\gainsm{+2.38} & \textbf{80.50}\,\gainsm{+2.34} & \textbf{83.19}\,\gainsm{+2.54} & \textbf{63.30}$^\dagger$\,\gainsm{+2.65} & \textbf{68.99}\,\gainsm{+2.92} & \textbf{77.60}\,\gainsm{+3.47} & \textbf{71.56}\,\gainsm{+0.28} & \textbf{79.78}\,\gainsm{+0.36} & \textbf{92.64}\,\gainsm{+0.44} & \textbf{70.42}\,\gainsm{+0.24} & \textbf{81.04}\,\gainsm{+0.21} & \textbf{91.43}\,\gainsm{+0.15}\\ %
\midrule
\multicolumn{13}{@{}l}{\emph{Reported SAM\,2.1-L results (other runs and scorers; AUC only):}}\\
SENTRY-D4S~\cite{sentry} (self-reported) & 76.3 & & & 61.8 & & & & & & & & \\ %
SAMITE~\cite{samite} (by~\cite{sam3rl}, official rule) & 75.8 & & & 61.9 & & & & & & & & \\
SAM2RL~\cite{sam2rl} (by~\cite{sam3rl}, official rule) & 75.4 & & & 61.1 & & & & & & & & \\
DAM4SAM~\cite{dam4sam} (self-reported) & 75.1 & & & 60.9 & & & & & & & & \\
HiM2SAM~\cite{him2sam} (self-reported, relaxed rule) & 75.1 & & & 62.8 & & & & & & & & \\
SAMURAI~\cite{samurai} (self-reported, relaxed rule) & 74.2 & & & 61.0 & & & & & & & & \\
SAMOSA~\cite{samosa} (self-reported, SAMURAI-style rule) & -- & & & 62.97 & & & & & & & & \\
\bottomrule
\end{tabular}}
\end{table*}

\emph{Decision rule of \#3.} Per benchmark, the system must exceed \emph{both} (i) D (same run) and (ii) SENTRY-D4S (separate run), each by more than the Holm-corrected multiple of its standard error~\cite{holm}. The pooled reading averages the three benchmarks with equal weights; it must exceed 2\,s.e.\ for both comparisons, with no benchmark below zero by more than 2\,s.e.

\begin{table}[t]
\centering
\caption{Pre-registered tests (PARMA, 350 training sequences): (i) vs D (same run); (ii) vs SENTRY-D4S (separate run). ``Mult./thr.'' is estimate/s.e.\ of the weaker of (i) and (ii) against its Holm threshold. Better/worse = sequences changing by more than 2 points. The pooled $n$ is the sum of the three benchmarks.}
\label{tab:confirm}
{\small%
\begin{tabular}{@{}lcccc@{}}
\toprule
Benchmark ($n$) & (i) & (ii) & Mult./thr. & Grade\\
\midrule
LaSOT-ext (150) & \up{+2.65}\se{0.66}, 44/11 & \up{+2.65}\se{0.68}, 44/12 & 3.91 / 2.4 & \textbf{exceeds both}\\ %
UAV123 (123) & \up{+0.28}\se{0.15}, 5/2 & \up{+0.28}\se{0.16}, 5/3 & 1.75 / 2.24 & positive, not est.\\
NFS (100) & \up{+0.24}\se{0.14}, 2/0 & \up{+0.15}\se{0.12}, 1/0 & 1.22 / 2.0 & positive, not est.\\
Pooled (373) & \up{+1.06}\se{0.23}, 4.57 s.e. & \up{+1.02}\se{0.23}, 4.35 s.e. & -- & \textbf{holds}\\
\bottomrule
\end{tabular}}
\end{table}

The registered conclusion (Table~\ref{tab:confirm}) is that, on LaSOT-ext, UAV123 and NFS, the tracker is above DAM4SAM and SENTRY-D4S overall, and on LaSOT-ext also individually. The pooled result is driven mainly by LaSOT-ext.

On UAV123 and NFS there appears to be little room for choosing. Among the outputs of the five archived baselines, picking the best cluster with ground truth would gain only +0.76 and +0.27. We have not computed this bound for our own branches on these benchmarks. The arbiter showed no measured drop on either, but we do not claim it is better there. %

On LaSOT-ext, the DAM4SAM (separate run) is the strongest re-run baseline, and the margin over it is +1.92\se{0.71}. That comparison spans two separate runs and was not part of the registered test.

\paragraph{LaSOT (descriptive).} PARMA $-$ D (same run) is +2.38\se{0.45}, with 43 sequences better and 6 worse by more than 2 points; on the 235 sequences outside V it is +1.73\se{0.40}. %
A separate, fully online run on LaSOT, side by side with SENTRY-D4S in the same job, is reported in Sec.~\ref{sec:online}.
The comparison registered for LaSOT, PARMA (SENTRY-D4S-led) $-$ SENTRY-D4S, came out at +1.21\se{0.32} on all 280 sequences and +0.89\se{0.28} outside V, which meets its registered criterion. %
Because the main method was selected after this evaluation, we treat all LaSOT numbers of the main method as descriptive.

\paragraph{Empty outputs.}\label{app:empty} When the selected branch has no mask on a frame, the main convention writes an empty box and the scorer carries the previous reported box forward, which is the same treatment every baseline gets. The per-source alternative instead carries the selected branch's own last box.
\begin{itemize}
\item On LaSOT the two conventions differ by +0.03\se{0.03}. %
\item On LaSOT-ext, UAV123 and NFS they differ by at most 0.02 in every registered reading. The largest difference is for LaSOT-ext (ii), +2.65 vs +2.63. %
\item On TNL2K, reading (i) is +1.86 under the per-source convention and +1.89 under the main one; reading (ii) is +1.46 and +1.32, because the per-source convention lowers the conservative medoid more than it lowers PARMA. %
\end{itemize}

\section{TNL2K: protocol and further readings}
\label{app:tnl2k}

Every benchmark other than TNL2K had been used in some way during development (Appendix~\ref{app:devuse}). On them, learning beats the untrained medoid rule clearly only on LaSOT (2.5\,s.e.; just over 2\,s.e.\ on NFS and pooled, not significant on LaSOT-ext and UAV123; Table~\ref{tab:rules}), and the arbiter's main advantage, fewer harmed videos, might come simply from defaulting to D and switching rarely. We therefore asked two questions once, on TNL2K~\cite{tnl2k}, which the project had never used: does the arbiter still improve its own base tracker on data that played no part in the design, and does it beat a consensus rule that switches about as often?

\paragraph{Protocol (registered before any tracking job on TNL2K).}
\begin{itemize}
\item \textbf{System.} The frozen 350 version with the strictly causal recomputation, and the same container, code bundle and weights as for LaSOT-ext, UAV123 and NFS. Nothing was retrained or tuned. The job script differs from theirs only in the dataset and manifest fields; the scoring script was written for TNL2K, reviewed independently and frozen before any box of a TNL2K record was read.
\item \textbf{Data.} The 700 official test sequences, with ground truth and absent labels from the official toolkit. One sequence, Soccer\_video\_02, was excluded before any result existed, as the registration prescribes for sequences that cannot be packed: four of its frames in the official archive match the archive's checksums but are not images. The remaining 699 sequences were all scored; none failed.
\item \textbf{Conservative medoid.} The medoid reports the branch whose box agrees most with the other four (summed box IoU; two empty boxes count as agreeing). Its conservative version reports that branch only if its agreement exceeds D's by more than $\tau$, and D's box otherwise; $\tau = 0$ is the plain medoid. Like the arbiter, it defaults to D and needs a margin to switch. We fixed $\tau = 0.4$ before TNL2K by matching switching rates on the 583 training sequences: there the arbiter reports a box outside D's cluster on 6.55\% of frames, and the rule with $\tau = 0.4$ on 6.51\%. On those sequences the rule gains +2.17\se{0.36} over D, against +2.36\se{0.33} for the arbiter. The $\tau$ with the highest mean gain on the 350 training sequences, 0.1, switches almost as often as the plain medoid (6.91\% vs 7.25\% on the 583), so a comparison with it could not separate learning from conservativeness; we report it as an explanatory row.
\item \textbf{Readings.} (i) PARMA $-$ D (same run); (ii) PARMA $-$ the conservative medoid with $\tau = 0.4$ on the same record. Holm-corrected: the reading with the larger multiple must exceed 2.24\,s.e., the other 2.0\,s.e.
\end{itemize}

\begin{table*}[t]
\centering\small
\caption{External test on TNL2K (699 sequences, read-out \ReadU, official rule). All rules use the same branch record. Gain is over D (s.e.), except in the two rows of registered readings, which give the difference named in the row. Better/worse = sequences changing by more than 2 points. Worst 5\% = mean change of the 35 sequences that lose most. Switched = frames whose reported box lies outside D's cluster (mean over sequences). The bounds choose with ground truth.}
\label{tab:tnl2k}
\begin{tabular}{@{}lccccccc@{}}
\toprule
Rule & AUC & P@20 & NP & Gain over D & Better/worse & Worst 5\% & Switched\\
\midrule
D (same run) & 60.83 & 70.62 & 68.35 & 0 & -- & -- & 0\\ %
Medoid ($\tau = 0$) & 61.32 & 71.09 & 68.42 & \up{+0.49}\se{0.32} & 106/70 & \down{$-$21.12} & 7.68\%\\
Conservative medoid, $\tau = 0.1$ & 61.41 & 71.20 & 68.54 & \up{+0.57}\se{0.30} & 100/67 & \down{$-$18.85} & 7.25\%\\
Conservative medoid, $\tau = 0.4$ (registered) & 61.40 & 71.15 & 68.64 & \up{+0.57}\se{0.29} & 87/57 & \down{$-$16.84} & 6.64\%\\
\textbf{PARMA} (350) & \textbf{62.73} & \textbf{72.43} & \textbf{70.05} & \textbf{\up{+1.89}\se{0.26}} & 134/22 & \down{$-$7.45} & 8.97\%\\
\midrule
\multicolumn{8}{@{}l}{\emph{Registered readings (Holm; thresholds 2.24 and 2.0 s.e.):}}\\
(i) PARMA $-$ D (same run) & & & & \up{+1.89}\se{0.26} & 134/22 & \multicolumn{2}{c}{7.20 s.e., \textbf{holds}}\\ %
(ii) PARMA $-$ conservative medoid, $\tau = 0.4$ & & & & \up{+1.32}\se{0.21} & 119/35 & \multicolumn{2}{c}{6.40 s.e., \textbf{holds}}\\
\midrule
Cluster-level bound of our five branches & & & & \up{+3.30} & & & \\ %
Branch-level bound of our five branches & & & & \up{+4.75} & & & \\
\bottomrule
\end{tabular}
\end{table*}

\paragraph{Registered result.} Both readings hold (Table~\ref{tab:tnl2k}). On a test set that played no part in the design, the arbiter significantly improves DAM4SAM, and it is significantly better than a consensus rule calibrated to its switching rate on training sequences.
\begin{itemize}
\item \textbf{(i)} PARMA $-$ D is +1.89\se{0.26}, 7.20\,s.e.; 134 sequences improve and 22 get worse by more than 2 points.
\item \textbf{(ii)} PARMA $-$ conservative medoid is +1.32\se{0.21}, 6.40\,s.e. The rule itself gains only +0.57\se{0.29} over D.
\end{itemize}
Our registered predictions had been +0.5 to +2.0 for (i) and +0.2 to +1.0 for (ii); (ii) came out above that range. %

\begin{figure}[t]
\centering
\includegraphics[width=0.55\textwidth]{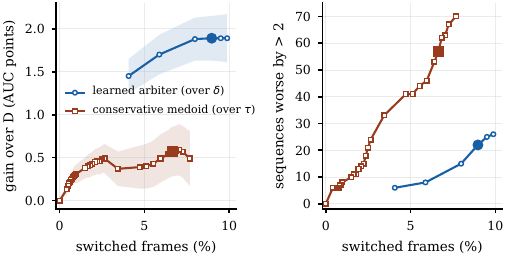}
\caption{TNL2K: gain over D (left, $\pm$1 s.e.) and sequences made worse by more than 2 points (right) against the share of switched frames, i.e.\ frames whose box lies outside D's cluster (as in Table~\ref{tab:tnl2k}). The arbiter is swept over its margin $\delta$ (0--0.3), the conservative medoid over $\tau$ (0--3.0). Filled markers are the registered configurations ($\delta = 0.05$, $\tau = 0.4$). Explanatory: the sweep was registered as a description, not a test.}
\label{fig:tnl2k}
\end{figure}

\paragraph{Explanatory readings (no pass line).}
\begin{itemize}
\item \textbf{Harm.} The arbiter's lower harm does not appear to come from conservativeness alone. With its switching rate matched on training data, the consensus rule still harms more than twice as many sequences (57 vs 22), and its worst 5\% of sequences change by $-$16.84 points on average, against $-$7.45 for the arbiter.
\item \textbf{Switching rate on TNL2K.} Because $\tau$ was matched on training data, the two systems do not switch equally often on TNL2K: the arbiter leaves D's cluster on 8.97\% of frames, the rule on 6.64\%. No $\tau$ makes the rule switch as often as the arbiter does at $\delta = 0.05$ (the plain medoid reaches 7.68\%). At $\delta = 0.2$ the arbiter switches on 5.88\% of frames and still gains +1.70\se{0.23}, with 8 sequences worse; the rule interpolated to the same rate gains +0.48, with about 46 worse (Fig.~\ref{fig:tnl2k}). This alignment was made after seeing TNL2K.
\item \textbf{Other untrained rules.} PARMA exceeds the plain medoid by +1.40\se{0.23} and the conservative medoid with $\tau = 0.1$ by +1.32\se{0.22}.
\item \textbf{Other conventions and measures.} Under the per-source empty-frame convention, under read-out U+1, and with P@20 or NP instead of AUC, (i) lies between +1.70 and +1.88 and (ii) between +1.28 and +1.46, each at least 5.0\,s.e.
\item \textbf{Headroom.} With ground truth choosing per frame, the best cluster would add +3.30 over D and the best branch +4.75; the arbiter obtains 57\% of the cluster-level bound. Counting a frame as a failure when its overlap is at most 0.5, each branch fails on 25.2--27.3\% of frames, and the pairwise correlations ($\phi$) of these failures are 0.67--0.87, so many failures are shared.
\item \textbf{Absolute numbers and speed.} PARMA reaches 62.73 AUC and D 60.83. The recording jobs ran at a median of 0.154\,s per frame on a mix of L40, L40S and H100 GPUs (0.139\,s on L40S), without the arbiter.
\end{itemize}

\paragraph{TNL2K scoring details.} About 14{,}500 TNL2K frames have a ground-truth box with $x \leq 0$ or $y \leq 0$; under the official rule they count as failures for AUC and as hits for P@20 and NP, for every system alike. Our scorer uses the overlap thresholds $k/20$; with MATLAB's \texttt{0:0.05:1} thresholds, four of which differ from $k/20$ in floating point, no sequence AUC changes by more than 0.05 points and both registered readings are unchanged. %

\section{Online run on LaSOT}
\label{app:online}
\label{sec:online}

The results in the main text were scored by applying the frozen arbiter to online branch records. We also ran the complete tracker online on all 280 LaSOT test sequences, one job per sequence. Each job ran the six branches and the arbiter, and ran SENTRY-D4S side by side in the same job. This gives a same-job comparison with SENTRY-D4S and removes the run-to-run variation that affects comparisons with archived baselines. The reading was registered before the run (Appendix~\ref{app:registration}, \#9). As for all LaSOT results, it is descriptive with respect to the choice of the main method.

\begin{table}[t]
\centering
\caption{Online run on LaSOT, all components in the same job (success AUC $\times 100$, paired differences). Better/worse = sequences changing by more than 2 points.}
\label{tab:online}
{\small%
\begin{tabular}{@{}lcc@{}}
\toprule
Comparison & All 280 & 235 outside V\\
\midrule
PARMA (SENTRY-D4S-led) $-$ SENTRY-D4S (registered primary) & \up{+1.70}\se{0.37}, 42/7 & \up{+1.30}\se{0.35}\\ %
PARMA $-$ D of the same job & \up{+1.82}\se{0.39}, 43/7 & \up{+1.43}\se{0.37}\\ %
PARMA $-$ SENTRY-D4S of the same job & \up{+1.61}\se{0.38}, 42/10 & \up{+1.21}\se{0.36}\\ %
D of the same job $-$ DAM4SAM (separate run) (two runs) & \up{+0.48}\se{0.29}, 24/11 & \up{+0.32}\se{0.27}\\ %
\bottomrule
\end{tabular}}
\end{table}

\begin{itemize}
\item \textbf{Primary reading.} The registered primary reading, PARMA (SENTRY-D4S-led) $-$ SENTRY-D4S of the same job, is +1.70\se{0.37} (4.57 s.e.). It is +1.30\se{0.35} on the 235 sequences outside V, so it meets its registered criterion (Table~\ref{tab:online}).
\item \textbf{Main method.} PARMA is +1.82\se{0.39} over D of the same job and +1.61\se{0.38} over SENTRY-D4S of the same job.
\item \textbf{Online equals offline.} The online decisions equal the offline, strictly causal recomputation on every frame (0 differing frames). The SENTRY-D4S boxes seen by the arbiter equal SENTRY-D4S's own output.
\item \textbf{Run-to-run variation.} D of this job differs from DAM4SAM (separate run) by +0.48\se{0.29}. This is the run-to-run variation that the same-job comparison removes.
\item \textbf{Speed.} The jobs ran at 0.646\,s per frame of wall time. This covers two processes side by side (SENTRY-D4S and the branch process with the arbiter), so it is not the speed of PARMA alone (Sec.~\ref{sec:speed}).
\end{itemize}

\section{Implementation checks and other benchmarks}
\label{app:branchdetails}
\paragraph{Implementation checks.}
On the training sequence crocodile-18, the branch records were byte-identical across a standalone run, a run alongside another tracker, and a run on a second L40S. In same-job checks on LaSOT racing-15 and yoyo-17, D reproduced standalone DAM4SAM and N reproduced DAM4SAM with promotion disabled. A100 and L40S outputs differed by 1--2 pixels on 44 of 1{,}000 frames. %

We do not evaluate on DiDi~\cite{dam4sam}: its protocol initialises and scores with masks, while our runner is box-initialised, and 86 of its 180 sequences are LaSOT test sequences. One TNL2K sequence was excluded before any result existed (Appendix~\ref{app:tnl2k}). %

\section{Ablations, speed and memory}
\label{app:ablation}

\begin{figure*}[t]
\centering
\includegraphics[width=\textwidth]{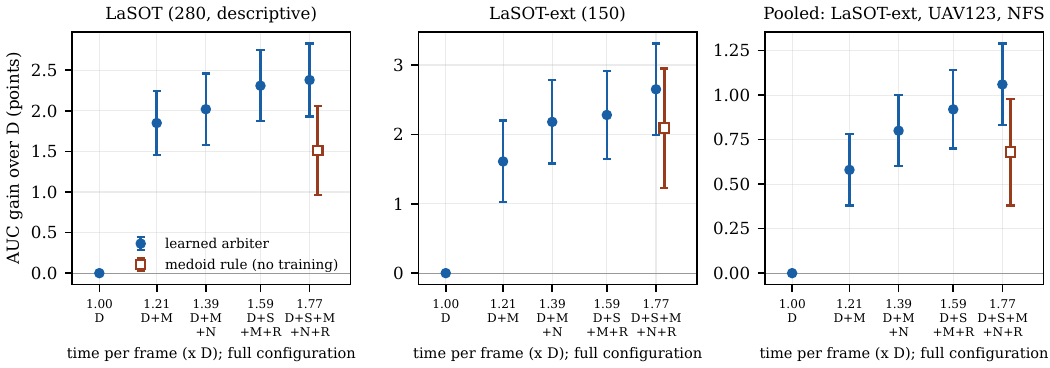}
\caption{Speed--accuracy trade-off on the test sets. The $y$-axis is the AUC gain over D (same run) ($\pm$1 s.e.). The $x$-axis is the time per frame relative to D alone, derived from per-branch timings on L40S, excluding record writing. Each point is a full configuration, labelled below the axis; D+S+M+N+R is the main system. The open square is the untrained medoid rule, which needs all five candidates. LaSOT is descriptive; the pooled panel is the pre-registered combination. Data: Tables~\ref{tab:branchabl} and~\ref{tab:rules}.}
\label{fig:speed}
\end{figure*}

We ablate on two disjoint sets:
\begin{itemize}
\item \textbf{Training sequences.} We train on the same 350 sequences and evaluate on 583 other LaSOT training sequences. These results inform design.
\item \textbf{Test set.} This ablation was registered before it was run and is reported in full regardless of outcome. %
\end{itemize}
All rows use the same recipe; only the inputs change. Fig.~\ref{fig:speed} plots the branch subsets against their time per frame.

\begin{table*}[t]
\centering
\caption{Learned arbiter vs rules without training: gain over D (s.e.). Better/worse = sequences changing by more than 2 points. The bound row uses ground truth.}
\label{tab:rules}
\resizebox{\textwidth}{!}{%
\begin{tabular}{@{}lccccccc@{}}
\toprule
Rule & Train (583) & Test LaSOT & Test LaSOT 235 & Test LaSOT-ext & Test UAV123 & Test NFS & Test pooled (3)\\
\midrule
Majority cluster & \up{+1.91}\se{0.34}, 83/32 & \up{+1.36}\se{0.47}, 36/18 & \up{+1.08}\se{0.46} & \up{+2.23}\se{0.79} & \up{+0.02}\se{0.11} & \down{$-$0.16}\se{0.19} & \up{+0.70}\se{0.27}\\ %
Highest predicted IoU & \up{+1.44}\se{0.35}, 103/49 & \up{+1.25}\se{0.51}, 40/31 & \up{+1.01}\se{0.48} & \down{$-$0.19}\se{0.60} & \up{+0.30}\se{0.16} & \up{+0.16}\se{0.17} & \up{+0.09}\se{0.21}\\
D unless two others agree against it & \up{+1.62}\se{0.43}, 95/54 & -- & -- & -- & -- & -- & --\\
R as the output & \up{+1.11}\se{0.29}, 73/30 & -- & -- & -- & -- & -- & --\\
Medoid (most agreed-with branch) & \up{+2.39}\se{0.39}, 123/56 & \up{+1.51}\se{0.55}, 49/26 & \up{+1.20}\se{0.55} & \up{+2.09}\se{0.86} & \up{+0.17}\se{0.17} & \down{$-$0.23}\se{0.21} & \up{+0.68}\se{0.30}\\
\textbf{PARMA (learned arbiter)} & \up{+2.36}\se{0.33}, 102/16 & \textbf{\up{+2.38}\se{0.45}, 43/6} & \textbf{\up{+1.73}\se{0.40}} & \textbf{\up{+2.65}\se{0.66}} & \up{+0.28}\se{0.15} & \textbf{\up{+0.24}\se{0.14}} & \textbf{\up{+1.06}\se{0.23}}\\
\midrule
PARMA $-$ medoid & \down{$-$0.03}\se{0.20} & \up{+0.87}\se{0.35} & \up{+0.53}\se{0.33} & \up{+0.56}\se{0.48} & \up{+0.11}\se{0.16} & \up{+0.47}\se{0.23} & \up{+0.38}\se{0.19}\\
Cluster-level bound of our five branches & \up{+3.95}\se{0.43} & -- & -- & -- & -- & -- & --\\
\bottomrule
\end{tabular}}
\end{table*}

\paragraph{Learned vs untrained rules (Table~\ref{tab:rules}).} The strongest untrained rule is the \emph{medoid}: report the branch whose box agrees most with the others.
\begin{itemize}
\item \textbf{Training sequences.} On the 583 sequences the medoid matches the learned arbiter's average gain (+2.39 vs +2.36), but it harms many more videos: 56 versus 16 drop by more than 2 points. On the 350 training sequences, where the arbiter is scored by class-grouped cross-validation, the arbiter is ahead by +0.50\se{0.32}. %
\item \textbf{Test set.} Learning adds +0.87\se{0.35} over the medoid on LaSOT, 2.5 s.e. The difference is not significant on the 235 sequences outside V (+0.53\se{0.33}) or on LaSOT-ext (+0.56\se{0.48}); on NFS (+0.47\se{0.23}) and pooled over LaSOT-ext, UAV123 and NFS (+0.38\se{0.19}) it is just above 2 s.e.\ (2.06 and 2.04). The medoid harms 26 LaSOT test sequences by more than 2 points; the arbiter harms 6.
\item \textbf{TNL2K (external test, Sec.~\ref{sec:tnl2k}).} Learning adds +1.40\se{0.23} over the plain medoid and, in the registered reading, +1.32\se{0.21} over a conservative medoid whose switching rate was matched on training data. %
\end{itemize}
We read this as follows. On the benchmarks used during development, a large part of the gain comes from running diverse memory policies and trusting their consensus, and learning adds a further, smaller gain, clearly on LaSOT and marginally on NFS and pooled. On the external TNL2K test, learning is clearly better than consensus, including a consensus rule calibrated to the arbiter's switching rate on training sequences, and it harms far fewer videos.

The rule that trusts the model's own confidence, i.e.\ picks the highest predicted IoU, is the weakest rule on LaSOT, on LaSOT-ext and pooled. On LaSOT-ext it falls below D, at $-$0.19\se{0.60}. It is not uniformly weakest: on UAV123 it is the best row (+0.30\se{0.16}), and on training sequences ``R as the output'' is weaker.

\begin{table}[t]
\centering
\caption{Amount of training data (training sequences; evaluation on the same 583). For 90 and 175, the first value uses the first sequences in the pool and the next two use random draws.}
\label{tab:size}
{\small%
\begin{tabular}{@{}lcc@{}}
\toprule
Training sequences & Gain over D & Share of bound\\
\midrule
90 (first; two random draws) & \up{+2.21}; \up{+2.32}, \up{+2.30} & 56--59\%\\ %
175 (first; two random draws) & \up{+2.13}; \up{+2.35}, \up{+2.27} & 54--59\%\\
350 (main) & \up{+2.36}\se{0.33} & 60\%\\
\bottomrule
\end{tabular}}
\end{table}

\paragraph{Training data (Table~\ref{tab:size}).} The arbiter needs few training videos. 350 sequences minus the mean of the three 90-sequence sets is +0.08\se{0.06}, and minus the mean of the three 175-sequence sets it is +0.11\se{0.05}.

\paragraph{Training on all 1{,}120 sequences.} As an illustration we also trained the arbiter on all 1{,}120 LaSOT training sequences (the ``full version''); it was not used for any decision. In class-grouped 10-fold cross-validation over the remaining 770 training sequences, adding them to training gives +2.21\se{0.29} instead of +2.06\se{0.28}, a paired difference of +0.16\se{0.09}. On the test sets its AUC is within 0.13 points of the 350 version (Table~\ref{tab:full}). %

\begin{table}[h]
\centering\small
\caption{350 vs 1{,}120 training sequences: AUC / $P_\text{norm}$ / $P$ on the test sets.}
\label{tab:full}
\begin{tabular}{@{}lcccc@{}}
\toprule
Training sequences & LaSOT & LaSOT-ext & UAV123 & NFS\\
\midrule
350 (main) & 77.20 / 80.50 / 83.19 & 63.30 / 68.99 / 77.60 & 71.56 / 79.78 / 92.64 & 70.42 / 81.04 / 91.43\\ %
1{,}120 (full) & 77.23 / 80.53 / 83.18 & 63.43 / 69.12 / 77.89 & 71.54 / 79.75 / 92.64 & 70.32 / 80.93 / 91.34\\
\bottomrule
\end{tabular}
\end{table}

\paragraph{Speed.}\label{sec:speed}
\begin{table}[t]
\centering
\caption{Seconds per frame: median over sequences of elapsed time divided by frames, from the run metadata of the 653 test sequences of LaSOT, LaSOT-ext, UAV123 and NFS. PARMA includes writing the branch records (0.008\,s on L40S) but not the arbiter.}
\label{tab:speed}
{\small%
\begin{tabular}{@{}lcccccc@{}}
\toprule
GPU & SAM\,2.1 & SAMURAI & HiM2SAM & DAM4SAM & SENTRY-D4S & PARMA\\
\midrule
L40S & 0.065 & 0.070 & 0.078 & 0.083 & 0.510 & 0.141\\ %
H200 & 0.062 & 0.066 & 0.078 & 0.062 & 0.517 & 0.106\\
H100 80GB & 0.059 & 0.063 & 0.078 & 0.063 & 0.494 & 0.104\\
A100 80GB & 0.078 & 0.085 & 0.095 & 0.081 & 0.699 & 0.153\\
\bottomrule
\end{tabular}}
\end{table}

\paragraph{Relative cost.} Measured directly within the same job, the full system costs about $2\times$ the time of D alone. Six runs, each with one extra diagnostic branch, gave 1.89--2.08$\times$ on L40S (four runs) and 2.16--2.21$\times$ on an RTX 4090 (two runs).

A finer breakdown comes from subtracting per-branch timings within our own runs, which avoids measuring each subset separately. It gives $1.77\times$ on L40S, or $1.88\times$ including record writing. Across the GPUs measured, this derived ratio is 1.77--1.89$\times$ excluding and 1.9--2.1$\times$ including record writing. %

\paragraph{Absolute speed.} On L40S, the jobs that produced our records ran at about 7 frames per second; this includes record writing but not the arbiter. That is about $3.6\times$ faster than SENTRY-D4S in its public default configuration (0.141 vs 0.510\,s per frame).

\paragraph{Caveats.}
\begin{itemize}
\item These timings come from separate runs, not from one machine at one time, and they mix torch versions (2.1.0 to 2.7.1).
\item Whether frame decoding is inside the baselines' per-frame times cannot be told from their metadata.
\item The SENTRY paper reports 30.2\,fps for SENTRY-D4S on an A100. We could not reproduce this with the public default configuration and have not resolved the discrepancy. %
\end{itemize}

\paragraph{Memory.} We measured peak allocated GPU memory (torch) on four LaSOT test sequences of 1{,}000 to 9{,}999 frames, running D alone and the six-branch system one after the other in fresh processes on the same GPU (Table~\ref{tab:memory}). The six branches add 23--28\,MiB to D alone, 0.25--1.07\%, and D's boxes were identical in both runs. The SENTRY-D4S column is the peak recorded in the online LaSOT jobs (Appendix~\ref{app:online}), with the same container and torch version. %

\begin{table}[h]
\centering\small
\caption{Peak allocated GPU memory (GiB). GPU: L40S, except spider-20 (L40).}
\label{tab:memory}
\begin{tabular}{@{}lcccc@{}}
\toprule
Sequence & Frames & D alone & PARMA (six branches) & SENTRY-D4S\\
\midrule
guitar-16 & 1{,}000 & 2.47 & 2.49 & 4.41\\ %
basketball-7 & 2{,}101 & 3.27 & 3.30 & 5.21\\
bird-17 & 5{,}000 & 5.40 & 5.43 & 7.34\\
spider-20 & 9{,}999 & 9.07 & 9.09 & 11.01\\
\bottomrule
\end{tabular}
\end{table}

\paragraph{Retraining with other seeds.} In the final fit, the regressor's random seed only decides which 10\% of the training rows scikit-learn holds out for its early-stopping check; every fit still ran the full 300 iterations. We refitted the frozen configuration with seeds 1--4 instead of 0, changing nothing else, and re-scored every test sequence of the five benchmarks with the same records and scorer (registration \#11). Across the five seeds (mean $\pm$ s.d.), mean AUC is 77.17$\pm$0.02 on LaSOT, 63.28$\pm$0.05 on LaSOT-ext, 71.56$\pm$0.01 on UAV123, 70.40$\pm$0.01 on NFS and 62.72$\pm$0.01 on TNL2K. On these five benchmarks, no seed is more than 0.10 points from the reported result (seed 0). For every seed, PARMA $-$ D is positive on all five benchmarks; as for seed 0, it exceeds 2 s.e.\ on LaSOT, LaSOT-ext and TNL2K and stays below 2 s.e.\ on UAV123 and NFS. TNL2K reading (ii) also exceeds 2 s.e.\ for every seed, and on each of the four benchmarks every seed stays above the five re-run baselines and D in the upper block of Table~\ref{tab:main}. %

\section{When and why the arbiter switches: summary}
\label{app:interp}
\begin{figure}[h]
\centering
\includegraphics[width=\textwidth]{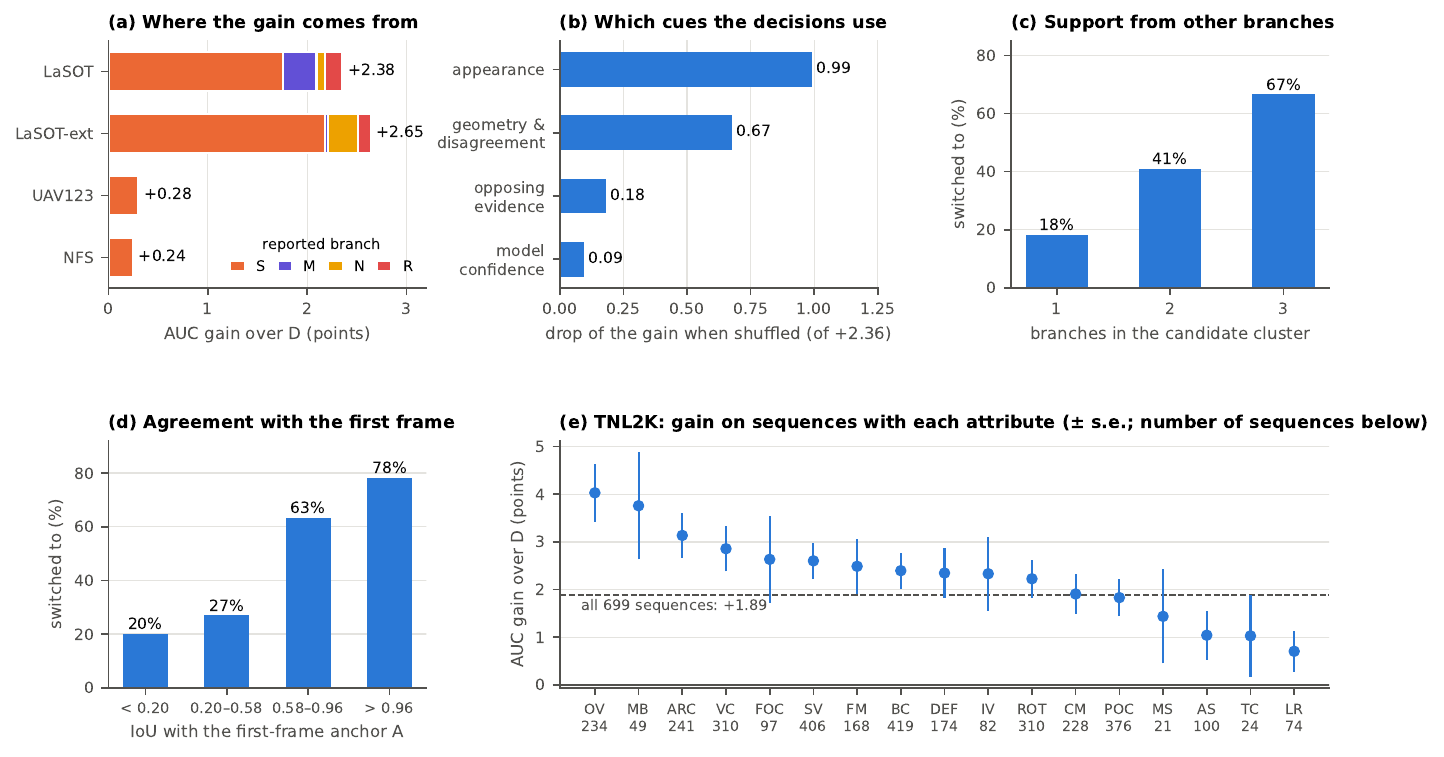}
\caption{(a) Gain over D on the four test benchmarks, split by the branch reported. (b) Drop of the +2.36 gain on 583 held-out training sequences when one cue group is shuffled. (c) Share of candidate clusters the arbiter switches to, by the number of branches in the cluster. (d) The same by the cluster's IoU with the first-frame anchor A. (e) TNL2K: gain over D on sequences with each attribute ($\pm$1 s.e.; number of sequences below each label).}
\label{fig:interp}
\end{figure}

Fig.~\ref{fig:interp} summarises the analyses of Sec.~\ref{sec:why}. They are descriptive: they read existing records and scored outputs and change no registered reading. Panels (a) and (e) use test sequences; panels (b)--(d) use the 583 held-out training sequences. The depth-3 surrogate tree was fitted on half of those sequences and checked on the other half: it agrees with the arbiter on 81\% of rows and keeps +2.04 of the arbiter's +2.13 on that half (96\%). %

\section{Box read-out, published numbers, and why they differ}
\label{app:readout}

\paragraph{Code.} Table~\ref{tab:code} lists how each official code base turns a mask into a box (we checked the latest public commit of each repository on 2026-11-22). No code base adds a constant beyond the pixel count. Under the LaSOT toolkit's convention, where \texttt{calc\_rect\_int.m} sets right $= x+w-1$, the pixel-count box \ReadU{} is the one aligned with the mask.

\begin{table}[h]
\centering\small
\caption{Mask-to-box rule in official code.}
\label{tab:code}
\begin{tabular}{@{}l p{0.26\textwidth} l p{0.36\textwidth}@{}}
\toprule
System & Width written & Equals & Scorer in its repository\\
\midrule
SAMURAI & $x_{\max}-x_{\min}$ & \ReadU$-1$ & own \texttt{lib/test/analysis}; valid frames need only $w,h>0$\\
HiM2SAM & $x_{\max}-x_{\min}$ & \ReadU$-1$ & same as SAMURAI's\\
DAM4SAM & $x_{\max}-x_{\min}+1$ & \ReadU & not specified\\
SENTRY & $x_{\max}-x_{\min}+1$ & \ReadU & none in the repository\\
SAM2Long & no box-tracking code in its repository & -- & --\\
SAM3RL & ``tight box''; code not released & ? & PyTracking (official rule)\\
LaSOT toolkit / PyTracking & -- & \ReadU{} is aligned & valid frames need all four ground-truth values $>0$\\
\bottomrule
\end{tabular}
\end{table}

\paragraph{Read-out U+1.} U+1 adds one pixel to the width and height on every frame with a box. The constant was first estimated in a preliminary experiment of ours on LaSOT training sequences. We report it only for comparison with published numbers. %

\begin{table}[h]
\centering\small
\caption{LaSOT and LaSOT-ext AUC under the official valid-frame rule: published numbers vs our re-runs in \ReadU{} and U+1. Published numbers are from the original papers.}
\label{tab:published}
\begin{tabular}{@{}lcccccc@{}}
\toprule
System & LaSOT published & LaSOT \ReadU & LaSOT U+1 & LaSOT-ext published & LaSOT-ext \ReadU & LaSOT-ext U+1\\
\midrule
SAM\,2.1 & 70.0 (as reported in~\cite{sam2long,dam4sam}) & 69.49 & 70.07 & 56.9 (as reported in~\cite{sam2long}) & 57.24 & 56.59\\ %
SAMURAI & 74.2 & 74.88 & 75.54 & 61.0 & 59.96 & 59.18\\
HiM2SAM & 75.1 & 74.46 & 75.12 & 62.8 & 60.42 & 59.64\\
DAM4SAM & 75.1 & 74.78 & 75.50 & 60.9 & 61.38 & 60.63\\
SENTRY-D4S & 76.3 & 75.58 & 76.31 & 61.8 & 60.65 & 59.92\\
PARMA (350) & -- & 77.20 & 77.94 & -- & 63.30 & 62.49\\
\bottomrule
\end{tabular}
\end{table}

\begin{table}[h]
\centering\small
\caption{All systems in U+1 (success AUC, official rule).}
\label{tab:uplus1}
\begin{tabular}{@{}lcccc@{}}
\toprule
Method & LaSOT & LaSOT-ext & UAV123 & NFS$^\dagger$\\
\midrule
SAM\,2.1 & 70.07 & 56.59 & 68.79 & 70.32\\ %
SAMURAI & 75.54 & 59.18 & 71.29 & 71.04\\
HiM2SAM & 75.12 & 59.64 & 70.51 & 67.93\\
DAM4SAM & 75.50 & 60.63 & 71.43 & 71.19\\
SENTRY-D4S & 76.31 & 59.92 & 71.50 & 71.20\\
D (same run) & 75.55 & 59.93 & 71.50 & 71.11\\
PARMA (350) & 77.94 & 62.49 & 71.79 & 71.33\\
PARMA (1{,}120 training sequences) & 77.97 & 62.60 & 71.77 & 71.24\\
\bottomrule
\end{tabular}
\end{table}

\paragraph{Observations on the read-out.}
\begin{enumerate}
\item \textbf{LaSOT.} U+1 brings HiM2SAM (75.12 vs 75.1), SENTRY-D4S (76.31 vs 76.3) and SAM\,2.1 (70.07 vs 70.0) close to their published values. This is a numerical match only, since no code base uses U+1. For DAM4SAM, \ReadU{} is closer (74.78 vs 75.1) than U+1 (75.50). SAMURAI's self-reported 74.2 is below both readings, and SAMURAI's and HiM2SAM's code write boxes one pixel \emph{smaller} than \ReadU.
\item \textbf{LaSOT-ext.} U+1 lowers every system there, by 0.65--0.83 points. %
An analysis of ours on LaSOT-ext indicates that the ground-truth boxes of the many small targets in LaSOT-ext are drawn tighter than the mask's bounding rectangle, so the effect of the read-out depends on annotation habits.
\item \textbf{Conclusions are unchanged.} In U+1, the margin over the strongest baseline is +1.62\se{0.40} on LaSOT, +1.86\se{0.70} on LaSOT-ext, +0.28\se{0.17} on UAV123 and +0.13\se{0.11} on NFS; in \ReadU{} it is +1.62, +1.92, +0.22 and +0.15. In U+1 the strongest UAV123 baseline is SENTRY-D4S rather than SAMURAI. %
\end{enumerate}

\paragraph{Scorer: the valid-frame rule.} The scorer in SAMURAI's code base, which HiM2SAM reuses, counts frames whose ground-truth box has positive width and height but $x\leq 0$ or $y\leq 0$. The official toolkit and PyTracking treat these frames as failures. SAM3RL reports that such implementations inflate LaSOT-ext scores by up to 1.7 AUC~\cite{sam3rl}. We re-scored all systems with only this rule relaxed; the scorer is otherwise unchanged. %
\begin{itemize}
\item \textbf{LaSOT.} No scored frame of the 280 test sequences is affected, so every score is identical under both rules. This rule therefore does not explain the LaSOT differences in Table~\ref{tab:published}, and their cause remains unexplained. Those differences are also not uniform:
  \begin{itemize}
  \item SAMURAI's self-reported number is 0.7 below our \ReadU;
  \item SAM3RL's re-measurements of SAMURAI (74.8) and HiM2SAM (74.3) are close to \ReadU;
  \item the self-reported numbers of SENTRY-D4S and HiM2SAM, and SAM3RL's DAM4SAM (75.5), are close to U+1.
  \end{itemize}
\item \textbf{LaSOT-ext.} 6{,}826 frames, 1.97\% of scored frames in 61 sequences, are affected, and relaxing the rule raises every system by 1.44--1.60 points (Tables~\ref{tab:relaxbase} and~\ref{tab:relaxed}).
  \begin{itemize}
  \item SAM3RL's re-measurements under the official rule are within 0.04--0.34 of our official-rule \ReadU{} values.
  \item SAMURAI's self-reported 61.0 is close to the relaxed rule applied to its own boxes (61.27).
  \end{itemize}
  This rule therefore explains why SAMURAI and HiM2SAM report higher LaSOT-ext numbers than our re-runs. All results in the paper use the official rule.
\end{itemize}

\begin{table}[h]
\centering\small
\caption{LaSOT-ext baselines: published numbers, SAM3RL's re-run, and our re-runs under both rules.}
\label{tab:relaxbase}
\begin{tabular}{@{}lccccc@{}}
\toprule
System & Published & SAM3RL re-run (official) & \ReadU, official & \ReadU, relaxed & Own boxes, relaxed\\
\midrule
SAM\,2.1 & 56.9 (as reported in~\cite{sam2long}) & 56.9 & 57.24 & 58.69 & 58.69\\ %
SAMURAI & 61.0 & 60.0 & 59.96 & 61.45 & 61.27\\
HiM2SAM & 62.8 & 60.5 & 60.42 & 62.02 & 61.83\\
DAM4SAM & 60.9 & 61.7 & 61.38 & 62.95 & 62.95\\
SENTRY-D4S & 61.8 & -- & 60.65 & 62.21 & 62.21\\
\bottomrule
\end{tabular}
\end{table}

\begin{table}[h]
\centering\small
\caption{LaSOT-ext, all systems under the relaxed valid-frame rule (read-out \ReadU; registered \#8, explanatory only).}
\label{tab:relaxed}
\begin{tabular}{@{}lcccc@{}}
\toprule
System & Relaxed rule & Official rule & Relaxed $-$ official & PARMA $-$ system, relaxed\\
\midrule
SAM\,2.1 & 58.69 & 57.24 & \up{+1.44}\se{0.45} & \up{+6.19}\se{1.28}\\ %
SAMURAI & 61.45 & 59.96 & \up{+1.49}\se{0.46} & \up{+3.43}\se{0.98}\\
HiM2SAM & 62.02 & 60.42 & \up{+1.60}\se{0.47} & \up{+2.86}\se{0.88}\\
DAM4SAM & 62.95 & 61.38 & \up{+1.57}\se{0.47} & \textbf{\up{+1.93}\se{0.72}} (strongest)\\
SENTRY-D4S & 62.21 & 60.65 & \up{+1.56}\se{0.47} & \up{+2.66}\se{0.68}\\
D (same run) & 62.23 & 60.65 & \up{+1.58}\se{0.47} & --\\
PARMA (350) & 64.88 & 63.30 & \up{+1.58}\se{0.47} & --\\
PARMA (1{,}120 training sequences) & 64.97 & 63.43 & \up{+1.54}\se{0.47} & --\\
\bottomrule
\end{tabular}
\end{table}

The relaxed rule raises every system by about the same amount, so the differences between systems barely change: against the strongest baseline the margin is +1.93 under the relaxed rule and +1.92 under the official one.

\paragraph{Archived box files.} Reading SAMURAI and HiM2SAM from their own archived box files rather than from their masks lowers them by 1.08--1.23 points on LaSOT, UAV123 and NFS, and by 0.17--0.18 on LaSOT-ext. All tables in the paper read every system from its masks. %

\section{The 95 cues}
\label{app:cues}
``Own'' is the cluster's representative and ``opp'' its opponent (Sec.~\ref{sec:arbiter}). ``Onset'' is the first frame of the current disagreement. Unless stated otherwise, medians and means are over the previous 30 frames.

\paragraph{Geometry and disagreement structure (36).}
\begin{itemize}
\item Cluster structure: cluster size; number of clusters; whether it is D's cluster; size of the largest other cluster; share of the last 30 and of the last 150 frames in which own was in the largest cluster; membership indicators for D, S, M, N, R.
\item Track smoothness: own empty; IoU with own previous box, and its 10- and 30-frame means; IoU of the constant-velocity extrapolation with the current box; log-area minus its 30-frame median, and minus frame 0; log aspect ratio minus frame 0; relative box size; share of empty frames in the last 30; frames since onset (log, capped at 1{,}500).
\item Onset and drift: own and opp smoothness at onset (IoU with previous box, extrapolation IoU, emptiness); own and opp drift (IoU of the current box with the box just before onset).
\item Mask shape: mask fill of its tight box; number of connected components; share of the largest component; each minus its 30-frame median.
\end{itemize}

\paragraph{Model confidence (23).}
\begin{itemize}
\item Object score; predicted IoU of the chosen candidate; best predicted IoU among the rejected candidates; their margin; object score and predicted IoU minus their 30-frame medians.
\item Opp's object score, chosen predicted IoU and margin.
\item For own and for opp: mean and 10th percentile of the logits inside the mask, and the share of near-zero logits.
\item S's Kalman score and phase; frames since D and S last agreed; number of D's long-term memory frames; indicators of which branch represents the cluster.
\end{itemize}

\paragraph{Opposing evidence (15).}
\begin{itemize}
\item Mask IoU of own and opp: current value, and 10- and 30-frame means (3).
\item Four quantities from own's side: the largest IoU between opp's mask and one of own's rejected candidates; that candidate's predicted IoU; the mean of own's logits inside opp's mask; that mean over 30 frames.
\item The same four quantities from opp's side (4).
\item At onset: the two ``is the other's mask my rejected candidate'' quantities and the two ``my logit inside the other's mask'' quantities (4).
\end{itemize}

\paragraph{Appearance (21).}
\begin{itemize}
\item For the mask-pooled image feature (memory-independent) and for the object pointer, seven quantities each:
  \begin{itemize}
  \item cosine similarity to the first-frames template (mean of frames 1--5 of D), and own minus opp;
  \item cosine similarity to an exponential moving-average template (factor 0.98, updated only on single-cluster frames), and own minus opp;
  \item cosine similarity to own's vector just before onset, and own minus opp;
  \item cosine similarity between own and opp.
  \end{itemize}
\item Cosine similarity of the object pointer to the prompted frame's pointer, and own minus opp.
\item Memory-conditioned feature: similarity to the vector just before onset, and own minus opp.
\item Anchor A: IoU of A's last box with own, own minus opp, and frames since A last ran.
\end{itemize}

The 12-cue linear control uses: support difference; D's-cluster indicator; frames since onset; object score; own and opp predicted-IoU margins; the four opposing-evidence indicators; image-feature drift difference; and mask-fill change.

\paragraph{Definitions and thresholds.} Logit statistics are computed on the low-resolution logit map of the chosen candidate: the mean and the 10th percentile inside the mask, and the number of pixels with $|\text{logit}| < 1$ divided by the mask area (``near-zero logits''). Templates are compared by cosine similarity, and the moving-average template uses the factor 0.98. %

\section{Example sequences}
\label{app:examples}
Table~\ref{tab:examples} lists the largest gains and losses vs D (same run): the five largest of each on LaSOT and LaSOT-ext, the two largest of each on UAV123, and the two largest gains and the largest loss on NFS (TNL2K is not included), with the longest stretches in which the output was not D's box. Frame numbers start from 0. %

\begin{table}[h]
\centering\small
\caption{Largest gains and losses vs D (same run) (AUC $\times 100$).}
\label{tab:examples}
\begin{tabular}{@{}llrcrl@{}}
\toprule
Benchmark & Sequence & $\Delta$ AUC & PARMA / D & Frames not D & Longest stretches (branch)\\
\midrule
LaSOT & gorilla-4 & \up{+47.49} & 60.60 / 13.11 & 1{,}412 of 2{,}053 & 1665--2014 M, 650--986 S, 1080--1381 S\\
LaSOT & electricfan-20 & \up{+47.11} & 78.97 / 31.85 & 1{,}294 of 2{,}305 & 1188--1652 S, 1705--1879 S\\
LaSOT & sheep-3 & \up{+42.36} & 80.00 / 37.64 & 1{,}008 of 2{,}097 & 1609--1804 S, 610--769 S\\
LaSOT & goldfish-10 & \up{+37.34} & 48.07 / 10.73 & 3{,}046 of 5{,}602 & 4677--4948 M, 3831--3957 S\\
LaSOT & coin-6 & \up{+36.67} & 73.67 / 37.00 & 1{,}740 of 3{,}120 & 1924--2600 M, 1720--1922 M\\
LaSOT & elephant-16 & \down{$-$8.80} & 36.61 / 45.41 & 686 of 2{,}661 & 1227--1284 S, 1966--2004 S\\
LaSOT & monkey-9 & \down{$-$8.44} & 63.31 / 71.75 & 934 of 4{,}079 & 512--599 S, 1860--1899 M\\
LaSOT & bird-2 & \down{$-$5.82} & 80.98 / 86.80 & 290 of 4{,}101 & 3832--3999 S, 4005--4100 S\\
LaSOT & car-17 & \down{$-$4.84} & 66.00 / 70.84 & 301 of 2{,}251 & 1898--2062 R, 2133--2183 R\\
LaSOT & mouse-8 & \down{$-$4.32} & 70.56 / 74.88 & 171 of 1{,}275 & 827--851 S, 788--807 S\\
\midrule
LaSOT-ext & dancingshoe-8 & \up{+45.76} & 69.96 / 24.20 & 1{,}550 of 2{,}500 & 1112--1370 S, 2247--2499 S\\
LaSOT-ext & skatingshoe-5 & \up{+41.70} & 66.78 / 25.08 & 1{,}403 of 2{,}500 & 570--1130 S, 1149--1390 S\\
LaSOT-ext & footbag-9 & \up{+28.74} & 44.30 / 15.56 & 1{,}284 of 2{,}500 & 1579--1714 S, 1491--1549 S\\
LaSOT-ext & jianzi-1 & \up{+25.92} & 33.50 / 7.59 & 1{,}587 of 2{,}500 & 1615--1719 S, 2303--2391 S\\
LaSOT-ext & dancingshoe-6 & \up{+25.21} & 77.68 / 52.47 & 1{,}018 of 2{,}500 & 1037--1194 N, 1561--1656 N\\
LaSOT-ext & dancingshoe-4 & \down{$-$22.34} & 44.86 / 67.20 & 931 of 2{,}500 & 458--565 S, 1472--1522 S\\
LaSOT-ext & raccoon-5 & \down{$-$13.86} & 69.91 / 83.77 & 405 of 2{,}232 & 1092--1294 S, 1310--1349 S\\
LaSOT-ext & cosplay-8 & \down{$-$10.35} & 64.82 / 75.17 & 386 of 2{,}500 & 865--882 M, 1458--1469 M\\
LaSOT-ext & lantern-5 & \down{$-$9.09} & 45.16 / 54.26 & 932 of 2{,}500 & 2386--2447 S, 2450--2499 S\\
LaSOT-ext & skatingshoe-4 & \down{$-$7.60} & 77.35 / 84.95 & 418 of 2{,}500 & 2340--2356 R, 2450--2464 S\\
\midrule
UAV123 & person10 & \up{+12.78} & 55.90 / 43.12 & 165 of 1{,}021 & 903--1020 S\\
UAV123 & uav7 & \up{+11.12} & 41.43 / 30.31 & 116 of 373 & 267--299 S, 242--258 S\\
UAV123 & bike2 & \down{$-$3.50} & 6.22 / 9.71 & 426 of 553 & 165--414 M, 108--163 M\\
UAV123 & bird1\_1 & \down{$-$3.05} & 26.67 / 29.72 & 24 of 253 & short S stretches\\
\midrule
NFS & beach\_flipback\_person & \up{+11.48} & 69.79 / 58.31 & 19 of 61 & 44--60 S\\
NFS & car\_white & \up{+6.67} & 58.38 / 51.71 & 197 of 2{,}063 & 1942--2062 S, 1874--1906 S\\
NFS & drone & \down{$-$1.02} & 5.44 / 6.46 & 1 of 70 & 63 M\\
\bottomrule
\end{tabular}
\end{table}

Most of the largest losses come from stretches switched to S; the longest such stretch is 203 frames, on raccoon-5 (frames 1092--1294). Some losses come from stretches switched to M (cosplay-8) or R (car-17). The list records which branch was reported on which frames, not why, so we do not attribute causes here.

\section{PARMA (SENTRY-D4S-led)}
\label{app:sentry}
PARMA (SENTRY-D4S-led) reports SENTRY-D4S (separate run) unless three things hold: the arbiter leaves D's cluster, SENTRY-D4S agrees with D, and SENTRY-D4S disagrees with the chosen branch. The numbers in Table~\ref{tab:sentryled} combine, frame by frame under the frozen rule, SENTRY-D4S (separate run) output with our six-branch records. On LaSOT the variant was also run online as one system, side by side in the same job; see Sec.~\ref{sec:online} and Table~\ref{tab:online}. Its cost is the sum of both systems.

\begin{table}[h]
\centering\small
\caption{PARMA (SENTRY-D4S-led) $-$ SENTRY-D4S (separate run) (350 version, main empty-frame convention).}
\label{tab:sentryled}
\begin{tabular}{@{}lcc@{}}
\toprule
Benchmark & Mean AUC & Difference (s.e.), better/worse\\
\midrule
LaSOT 280 & -- & \up{+1.26}\se{0.32}, 30/5\\ %
LaSOT 235 outside V & -- & \up{+0.91}\se{0.28}, 19/4\\
LaSOT-ext & 62.86 & \up{+2.21}\se{0.59}, 40/10\\ %
UAV123 & 71.51 & \up{+0.22}\se{0.13}, 4/2\\
NFS & 70.44 & \up{+0.17}\se{0.12}, 1/0\\
\bottomrule
\end{tabular}
\end{table}

The registered LaSOT reading (\#1) used the per-source convention. It gave a mean AUC of 76.79, a difference of +1.21\se{0.32} with 27/5 better/worse, and +0.89\se{0.28} with 17/4 on the 235 outside V. %

On LaSOT, PARMA (SENTRY-D4S-led) $-$ PARMA is $-$0.38\se{0.21} (per-source convention, registered evaluation \#1). Making SENTRY-D4S the leader therefore does not improve on PARMA. It also requires running both systems, which takes more than four times as long per frame (Table~\ref{tab:speed}: 0.510 + 0.141\,s vs 0.141\,s on L40S).

\section{Further details}
\label{app:details}

\paragraph{Branch subsets.} The full ablation has 31 rows on the 583 training sequences: the 16 branch subsets, each with and without A, except D alone, which has only the with-A row.
\begin{itemize}
\item The best subset of each size is D+M (88\%), D+M+N (92\%), D+M+N+R (98\%) and the full set (100\%). %
\item D+M+N+R needs S to be run because of R, so it costs as much as the full set. The best subset that really runs only four branches is D+S+M+R (96\%).
\end{itemize}

\paragraph{Run-to-run variation.} Our record's D minus DAM4SAM (separate run), i.e.\ the same tracker run twice, is $-$0.73\se{0.70} on LaSOT-ext, +0.08\se{0.09} on UAV123 and $-$0.08\se{0.08} on NFS. %
This is why the registered test compares with D (same run) as well as with SENTRY-D4S (separate run).

\paragraph{Software.} We used SAM\,2.1-L weights and the DAM4SAM runtime. Our records used torch 2.1.0 (training records and LaSOT) and 2.5.1 (LaSOT-ext, UAV123, NFS and TNL2K). The arbiter uses scikit-learn's \texttt{HistGradientBoostingRegressor}; the scoring runs used scikit-learn 1.9.0, Python 3.14.6 and NumPy 2.4.6. %

\section{Further limitations}
\label{app:limits}
\begin{itemize}
\item \textbf{Training data.} The arbiter's parameters are fitted only on 350 LaSOT training sequences, though its boosting settings were also used in the earlier feasibility probe. LaSOT-ext shares no object class with the training sequences, and UAV123, NFS and TNL2K come from other sources. We have not analysed how the cue distributions shift between benchmarks; the small gains on UAV123 and NFS may reflect little headroom or a shift the arbiter does not handle.
\item \textbf{Untrained consensus is strong on the development benchmarks.} On LaSOT-ext and UAV123 learning is not significantly better than the medoid; on NFS and pooled it is better by just over 2\,s.e. The learned part is supported by its gains on LaSOT and on the external TNL2K test, and by fewer harmed videos.
\item \textbf{Comparisons with archived baselines span runs.} Re-running the same tracker changes individual sequences by up to about 19 points, so comparisons with separate runs have larger standard errors than those with D (same run). %
\item \textbf{Reproduced baselines differ from their papers.} On LaSOT, four of our five re-runs are below their published numbers by 0.32--0.72 AUC, and SAMURAI is above its paper; on LaSOT-ext, SAM\,2.1 and DAM4SAM are above theirs. The box read-out explains part of the differences, and a permissive valid-frame rule explains why SAMURAI and HiM2SAM report higher LaSOT-ext numbers, but the LaSOT differences remain unexplained (Appendix~\ref{app:readout}). SENTRY-D4S was run in the default configuration of its public code, which may differ from the configuration behind its published numbers.
\item \textbf{Online execution.} On LaSOT, the complete tracker was run online on all 280 test sequences, and its output matched the offline, strictly causal recomputation on every frame (Appendix~\ref{app:online}). On LaSOT-ext, UAV123, NFS and TNL2K, the branch records were produced online and the arbiter's decisions are strictly causal, but the reported outputs were obtained by applying the frozen arbiter to these records offline.
\end{itemize}


\begin{thebibliography}{99}\small

\bibitem{aaa} H.~Song, D.~Suehiro, and S.~Uchida. AAA: Adaptive aggregation of arbitrary online trackers with theoretical performance guarantee. \emph{arXiv:2009.09237}, 2020.

\bibitem{bofn} B.~Alawode, S.~Javed, A.~Mahmood, and J.~Matas. Predicting the best of {N} visual trackers. \emph{arXiv:2407.15707}, 2024.

\bibitem{chtc} Center for High Throughput Computing. Center for High Throughput Computing. 2006. doi:10.21231/GNT1-HW21.

\bibitem{cocolot} M.~Dunnhofer and C.~Micheloni. {CoCoLoT}: Combining complementary trackers in long-term visual tracking. In \emph{ICPR}, 2022. arXiv:2205.04261.

\bibitem{dam4sam} J.~Videnovi\'c, A.~Luke\v{z}i\v{c}, and M.~Kristan. A distractor-aware memory for visual object tracking with {SAM2}. In \emph{CVPR}, 2025. arXiv:2411.17576.

\bibitem{dtnet} K.~Song, W.~Zhang, R.~Song, and Y.~Li. Online decision based visual tracking via reinforcement learning. In \emph{NeurIPS}, 2020.

\bibitem{fedus2022switch} W.~Fedus, B.~Zoph, and N.~Shazeer. Switch {T}ransformers: Scaling to trillion parameter models with simple and efficient sparsity. \emph{Journal of Machine Learning Research}, 23(120):1--39, 2022. arXiv:2101.03961.

\bibitem{hdt} Y.~Qi, S.~Zhang, L.~Qin, H.~Yao, Q.~Huang, J.~Lim, and M.-H. Yang. Hedged deep tracking. In \emph{CVPR}, pages 4303--4311, 2016.

\bibitem{him2sam} R.~Chen, G.~Sun, Y.~Li, J.~Qin, and L.~Benini. {HiM2SAM}: Enhancing {SAM2} with hierarchical motion estimation and memory optimization towards long-term tracking. \emph{arXiv:2507.07603}, 2025.

\bibitem{hmmtxd} T.~Voj\'i\v{r}, J.~Matas, and J.~Noskov\'a. Online adaptive hidden {M}arkov model for multi-tracker fusion. \emph{arXiv:1504.06103}, 2015.

\bibitem{holm} S.~Holm. A simple sequentially rejective multiple test procedure. \emph{Scandinavian Journal of Statistics}, 6(2):65--70, 1979.

\bibitem{jacobs1991moe} R.~A. Jacobs, M.~I. Jordan, S.~J. Nowlan, and G.~E. Hinton. Adaptive mixtures of local experts. \emph{Neural Computation}, 3(1):79--87, 1991. doi:10.1162/neco.1991.3.1.79.

\bibitem{lasot} H.~Fan, L.~Lin, F.~Yang, P.~Chu, G.~Deng, S.~Yu, H.~Bai, Y.~Xu, C.~Liao, and H.~Ling. {LaSOT}: A high-quality benchmark for large-scale single object tracking. In \emph{CVPR}, 2019.

\bibitem{lasotext} H.~Fan, H.~Bai, L.~Lin, F.~Yang, P.~Chu, G.~Deng, S.~Yu, Harshit, M.~Huang, J.~Liu, Y.~Xu, C.~Liao, L.~Yuan, and H.~Ling. {LaSOT}: A high-quality large-scale single object tracking benchmark. \emph{International Journal of Computer Vision}, 129:439--461, 2021.

\bibitem{ltmu} K.~Dai, Y.~Zhang, D.~Wang, J.~Li, H.~Lu, and X.~Yang. High-performance long-term tracking with meta-updater. In \emph{CVPR}, 2020. arXiv:2004.00305.

\bibitem{mcct} N.~Wang, W.~Zhou, Q.~Tian, R.~Hong, M.~Wang, and H.~Li. Multi-cue correlation filters for robust visual tracking. In \emph{CVPR}, pages 4844--4853, 2018.

\bibitem{nfs} H.~Kiani~Galoogahi, A.~Fagg, C.~Huang, D.~Ramanan, and S.~Lucey. Need for speed: A benchmark for higher frame rate object tracking. In \emph{ICCV}, 2017.

\bibitem{ptav} H.~Fan and H.~Ling. Parallel tracking and verifying: A framework for real-time and high accuracy visual tracking. In \emph{ICCV}, 2017. arXiv:1708.00153.

\bibitem{rethinkmem} M.~Alansari, M.~Naseer, H.~Al~Marzouqi, N.~Werghi, and S.~Javed. Rethinking memory design in {SAM}-based visual object tracking. \emph{arXiv:2512.22624}, 2025.

\bibitem{sam2} N.~Ravi, V.~Gabeur, Y.-T. Hu, R.~Hu, C.~Ryali, T.~Ma, H.~Khedr, R.~R\"adle, C.~Rolland, L.~Gustafson, E.~Mintun, J.~Pan, K.~V. Alwala, N.~Carion, C.-Y. Wu, R.~Girshick, P.~Doll\'ar, and C.~Feichtenhofer. {SAM 2}: Segment anything in images and videos. \emph{arXiv:2408.00714}, 2024.

\bibitem{sam2long} S.~Ding, R.~Qian, X.~Dong, P.~Zhang, Y.~Zang, Y.~Cao, Y.~Guo, D.~Lin, and J.~Wang. {SAM2Long}: Enhancing {SAM 2} for long video segmentation with a training-free memory tree. In \emph{ICCV}, 2025. arXiv:2410.16268.

\bibitem{sam2rl} A.~Adamyan, T.~\v{C}\'i\v{z}ek, M.~Straka, K.~Janouskova, and M.~Schmid. {SAM2RL}: Towards reinforcement learning memory control in segment anything model 2. \emph{arXiv:2507.08548}, 2025.

\bibitem{sam3} N.~Carion, L.~Gustafson, Y.-T. Hu, S.~Debnath, R.~Hu, D.~Suris, C.~Ryali, K.~V. Alwala, H.~Khedr, A.~Huang, et~al. {SAM 3}: Segment anything with concepts. \emph{arXiv:2511.16719}, 2025.

\bibitem{sam3rl} T.~\v{C}\'i\v{z}ek, M.~Straka, I.~Volkov, M.~Kripner, E.~Maci\'c, J.~Matas, K.~Janouskova, and M.~Schmid. {SAM3RL}: Memory control via reinforcement learning for visual object tracking. \emph{arXiv:2610.19931}, 2026.

\bibitem{samosa} D.~Zhu, Y.~Wang, Y.~Liu, Y.~Tang, B.~Yu, J.~Lu, and J.~Zhou. Segment anything with motion, geometry, and semantic adaptation for complex nonlinear visual object tracking. \emph{arXiv:2605.22538}, 2026.

\bibitem{samite} Q.~Xu, L.~Zhu, C.~Liu, G.~Lin, C.~Long, Z.~Li, and R.~Zhao. {SAMITE}: Position prompted {SAM2} with calibrated memory for visual object tracking. \emph{arXiv:2507.21732}, 2025.

\bibitem{samurai} C.-Y. Yang, H.-W. Huang, W.~Chai, Z.~Jiang, and J.-N. Hwang. {SAMURAI}: Adapting segment anything model for zero-shot visual tracking with motion-aware memory. \emph{arXiv:2411.11922}, 2024.

\bibitem{sentry} M.~Alansari, Y.~Michael, H.~AlMarzouqi, M.~Naseer, N.~Werghi, and S.~Javed. {SENTRY}: {SAM2}-enhanced neighbor-aware and temporally reasoned memory for visual tracking. In \emph{ECCV}, 2026. arXiv:2606.24449.

\bibitem{shazeer2017moe} N.~Shazeer, A.~Mirhoseini, K.~Maziarz, A.~Davis, Q.~Le, G.~Hinton, and J.~Dean. Outrageously large neural networks: The sparsely-gated mixture-of-experts layer. In \emph{ICLR}, 2017. arXiv:1701.06538.

\bibitem{sklearn} F.~Pedregosa et~al. Scikit-learn: Machine learning in {P}ython. \emph{Journal of Machine Learning Research}, 12:2825--2830, 2011.

\bibitem{tbt} M.~Dunnhofer, N.~Martinel, and C.~Micheloni. Tracking-by-trackers with a distilled and reinforced model. In \emph{ACCV}, 2020. arXiv:2007.04108.

\bibitem{tnl2k} X.~Wang, X.~Shu, Z.~Zhang, B.~Jiang, Y.~Wang, Y.~Tian, and F.~Wu. Towards more flexible and accurate object tracking with natural language: Algorithms and benchmark. In \emph{CVPR}, 2021. arXiv:2103.16746.

\bibitem{uav123} M.~Mueller, N.~Smith, and B.~Ghanem. A benchmark and simulator for {UAV} tracking. In \emph{ECCV}, 2016.

\end{thebibliography}
\end{document}